\documentclass[a4paper,fleqn]{cas-dc}

\usepackage[numbers]{natbib}

\usepackage{lineno}
\usepackage{framed,multirow}
\usepackage{booktabs}
\usepackage{tabularx}
\usepackage[table,dvipsnames]{xcolor} % oppure [svgnames] o [x11names]
\usepackage{url}
\usepackage{comment}
\usepackage{subcaption}
\usepackage[most]{tcolorbox}
\usepackage{fontawesome5}

\usepackage{float}

\def\tsc#1{\csdef{#1}{\textsc{\lowercase{#1}}\xspace}}
\tsc{WGM}
\tsc{QE}
\tsc{EP}
\tsc{PMS}
\tsc{BEC}
\tsc{DE}
\newcommand{\HSEP}{0.12em}   % orizzontale (0.08–0.20em)
\newcommand{\VRAISE}{+0.23ex} % verticale   (-0.10ex..+0.20ex) -> positivo = più in alto
\newcommand{\SCALE}{0.62}    % scala       (0.56–0.66)

\newcommand{\medalicon}[2]{%
  \makebox[0pt][l]{%
    \hspace{\HSEP}%
    \textcolor{#1}{\raisebox{\VRAISE}{\scalebox{\SCALE}{\faMedal}}}%
  }%
  \hspace{0.78em}% spazio “parcheggio” per l’icona
}

\newcommand{\benchmarkacron}{PrACo++}
\newcommand{\benchmarkname}{\underline{Pr}ompt\--\underline{A}ware \underline{Co}unting++}

\newcommand{\datasetacron}{MUCCA}
\newcommand{\datasetname}{\underline{MU}lti\-\underline{C}ategory \underline{C}lass\--\underline{A}gnostic counting}

\newcommand{\gold}{\medalicon{Goldenrod}}
\newcommand{\silver}{\medalicon{Gray}}
\newcommand{\bronze}{\medalicon{Brown}}

\tcbset{
    imgbox/.style={
        enhanced,
        arc=4pt,         % arrotondamento degli angoli (valore in pt OK)
        boxrule=2pt,     % spessore cornice
        colback=white,   % sfondo interno
        colframe=black,  % colore cornice (verrà sovrascritto nel singolo box)
        left=0pt, right=0pt, top=0pt, bottom=0pt,
        boxsep=0pt
    }
}

\begin{document}
\let\WriteBookmarks\relax
\def\floatpagepagefraction{.7} % Era 1, troppo difficile da soddisfare
\def\textpagefraction{.05}      % Era .001
\def\topfraction{.9}
\shorttitle{Does it Really Count?}
\shortauthors{G. Pacini et~al.}

\title [mode = title]{Does it Really Count? Assessing Semantic Grounding in Text-Guided Class-Agnostic Counting}                      
%\tnotemark[1,2]

%\tnotetext[1]{This document is the results of the research
%   project funded by the National Science Foundation.}

%\tnotetext[2]{The second title footnote which is a longer text matter
%   to fill through the whole text width and overflow into
%   another line in the footnotes area of the first page.}

\author[1,2]{Giacomo Pacini}[orcid=0009-0007-7745-4456] %type=editor,
                        % auid=000,bioid=1,
\cormark[1]
\ead{giacomo.pacini@isti.cnr.it}
%\ead[url]{www.jkkrishnan.in}

\credit{Conceptualization, Data curation, Investigation, Methodology, Resources, Validation, Writing - Original Draft, Writing - Review and Editing}

%\address[1]{, Street 129, 1043 NX Amsterdam, The Netherlands}
\affiliation[1]{organization={Institute of Information Science and Technologies of the National Research Council (ISTI-CNR)},
                addressline={Via G. Moruzzi 1}, 
%               citysep={}, % Uncomment if no comma needed between city and postcode
                postcode={56124}, 
                city={Pisa},
                country={Italy}}

\author[1]{Luca Ciampi}[orcid=0000-0002-6985-0439]
\cormark[1]
\ead{luca.ciampi@isti.cnr.it}

\credit{Conceptualization, Data curation, Investigation, Methodology, Resources, Validation, Writing - Original Draft, Writing - Review and Editing}

\author[1]{Nicola Messina}[orcid=0000-0003-3011-2487]
\cormark[1]
\ead{nicola.messina@isti.cnr.it}

\credit{Conceptualization, Data curation, Investigation, Methodology, Resources, Validation, Writing - Original Draft, Writing - Review and Editing}

\author[2]{Nicola Tonellotto}[orcid=0000-0002-7427-1001]
\ead{nicola.tonellotto@unipi.it}

\credit{Funding Acquisition, Project Administration, Supervision, Writing – review and editing}

\author[2]{Giuseppe Amato}[orcid=0000-0003-0171-4315]
\ead{giuseppe.amato@isti.cnr.it}

\credit{Funding Acquisition, Project Administration, Supervision, Writing – review and editing}

\author[2]{Fabrizio Falchi}[orcid=0000-0001-6258-5313]
\ead{fabrizio.falchi@isti.cnr.it}

\credit{Funding Acquisition, Project Administration, Supervision, Writing – review and editing}

\affiliation[2]{organization={University of Pisa - Department of Information Engineering},
                addressline={Via G. Caruso 16}, 
                postcode={56122}, 
                postcodesep={}, 
                city={Pisa},
                country={Italy}}

\cortext[cor1]{Corresponding author}
%\cortext[cor2]{Principal corresponding author}
%\fntext[fn1]{This is the first author footnote, but is common to third
%  author as well.}
%\fntext[fn2]{Another author footnote, this is a very long %footnote and
%  it should be a really long footnote. But this footnote is not yet
%  sufficiently long enough to make two lines of footnote text.}

%\nonumnote{This note has no numbers. In this work we demonstrate $a_b$
%  the formation Y\_1 of a new type of polariton on the interface
%  between a cuprous oxide slab and a polystyrene micro-sphere placed
%  on the slab.
%  }

\begin{abstract}
Open-world text-guided class-agnostic counting (CAC) has emerged as a flexible paradigm for counting arbitrary object classes via natural-language prompts. However, current evaluation protocols primarily focus on standard counting errors within single-category images, overlooking a fundamental requirement: the ability to correctly ground the textual prompt in the visual scene. In this paper, we show that several state-of-the-art CAC models often struggle to determine \textit{which} object class to count based on the given prompt, revealing a misalignment between textual semantics and visual object representations. This limitation leads to spurious counting responses and reduced reliability in real-world scenarios. To systematically address these limitations, our contribution is two-fold: (i) we introduce \benchmarkacron{} (\benchmarkname{}), a novel test suite featuring two dedicated evaluation protocols---the negative-label test and the distractor test---paired with new specialized metrics; and (ii) we present the \datasetacron{} (\datasetname{}) evaluation dataset, a new collection of real-world images featuring multiple annotated object categories per scene, unlike existing CAC benchmarks that typically include a single category per image.
Our extensive experimental evaluation of 10 state-of-the-art methods shows that, despite strong performance under standard counting metrics, current models exhibit significant weaknesses in understanding and grounding object class descriptions.
%—limitations that are systematically revealed by our proposed benchmark and dataset. 
Finally, we provide a quantitative analysis of how semantic similarity between prompts influences these failures. Overall, our results underscore the need for more semantically grounded architectures and offer a reliable framework for future assessment in open-world text-guided CAC methods. The code for reproducing our results is available at \href{https://github.com/ciampluca/PrACo}{github.com/ciampluca/PrACo}.
\end{abstract}

%\includegraphics{figs/cas-grabs.pdf}

%\item Research highlights item 1
%\item Research highlights item 2
%\item Research highlights item 3


\maketitle

%\linenumbers
%\switchlinenumbers

\section{Introduction}
\label{sec:intro}
Class-agnostic counting (CAC) aims to count instances of arbitrary object classes beyond the categories encountered during training~\cite{DBLP:conf/cvpr/RanjanSNH21}. This recent paradigm addresses the inherent constraints of conventional class-specific counting approaches, which rely on dedicated models trained for pre-defined object types---e.g., vehicles~\cite{DBLP:conf/iccv/ZhangWCM17,DBLP:journals/eswa/CiampiGCFVA22,DBLP:conf/kst/SeenouvongWNKO16,8969620}, people~\cite{DBLP:conf/wacv/HossainHCW19,DBLP:journals/eswa/BenedettoCCFGA22,DBLP:conf/cvpr/LiuSF19,DBLP:journals/ivc/KhanMH23,DBLP:journals/ivc/ZhouH25,DBLP:journals/ivc/WangLQWZWW25,DBLP:journals/ivc/ZhouRLHS24}, or animals~\cite{DBLP:conf/eccv/ArtetaLZ16,DBLP:journals/pnas/NorouzzadehNKSP18,DBLP:journals/ecoi/CiampiZICBFAC23}. Unlike these methods, CAC allows users to dynamically define target categories during inference, removing the need to retrain deep learning networks with class-specific annotated datasets.

Target object classes in CAC can be specified by users either through visual exemplars---using bounding boxes highlighting sample objects within input images~\cite{DBLP:conf/bmvc/LiuZZX22,DBLP:journals/corr/abs-2405-11770,DBLP:conf/aaai/WangX0024,Dukic_2023_ICCV,DBLP:journals/tip/WuCLCWL25,DBLP:journals/spl/GongYZ25,DBLP:journals/ivc/XuLYWLZ25,DBLP:journals/ivc/ZhangZCWH25}---or text prompts containing natural language descriptions~\cite{DBLP:conf/aaai/0018C24,DBLP:conf/aaai/KangMKH24,10483595,DBLP:conf/mm/JiangLC23,DBLP:conf/cvpr/XuL0RS23,AminiNaieni23}. Both paradigms exhibit advantages and disadvantages~\cite{DBLP:journals/cviu/CiampiAASYEAF26}. On the one hand, exemplar-based methods typically perform better by providing rich visual context, such as appearance and spatial details. On the other hand, open-world text-guided approaches prioritize flexibility by reducing user input effort, as they do not require the provision of bounding boxes. Additionally, the latter techniques integrate effectively with popular vision-language foundation models, such as CLIP~\cite{DBLP:conf/icml/RadfordKHRGASAM21}, increasing their relevance and adoption in current research trends. For these reasons, this work focuses on the open-world text-guided CAC setting.

\begin{figure*}[!htbp]
    \centering
    \includegraphics[width=0.95\linewidth]{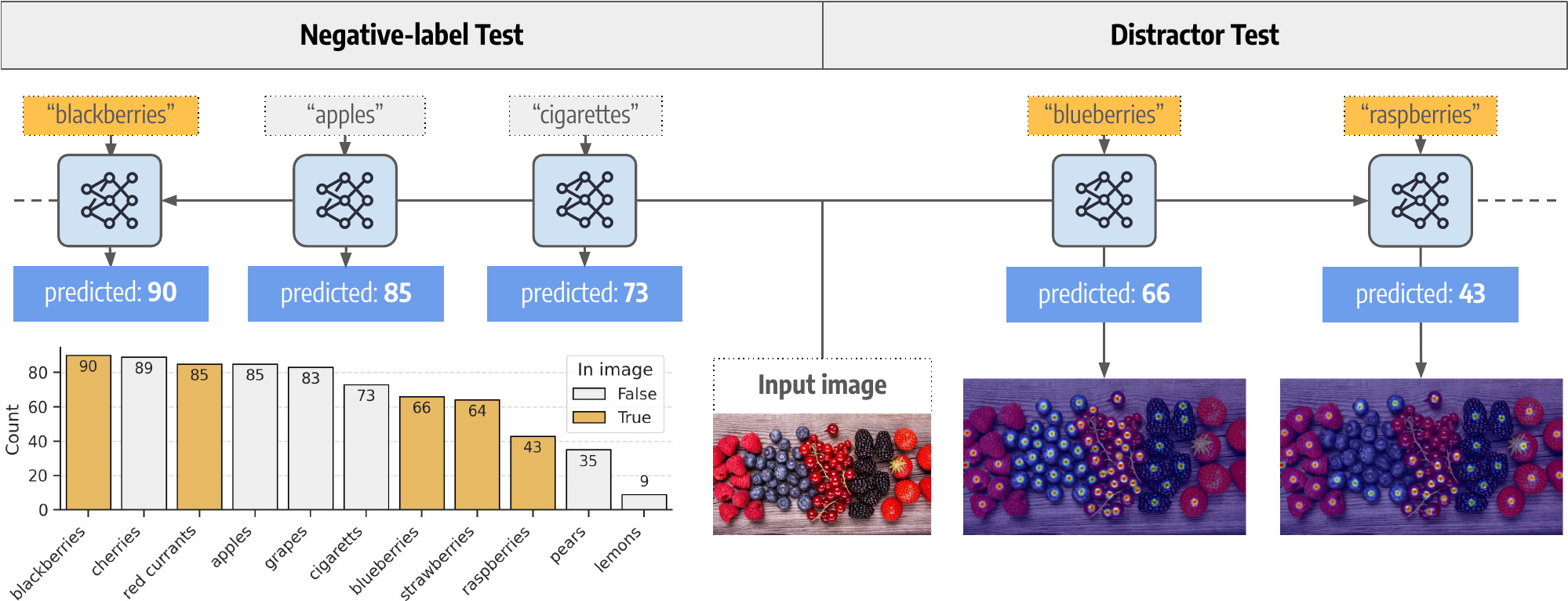}
    \caption{\textbf{High-level overview of our \benchmarkacron{} test suite.}
    We empirically show that SOTA open-world text-guided CAC methods are not properly evaluated by current benchmarks, as they fail to assess the alignment between textual semantic understanding and counting accuracy. To address this limitation, we introduce a new test suite, \benchmarkname{} (\benchmarkacron), composed of two complementary tests. (i) On the left, we illustrate the \textit{negative-label test}, which reveals cases where object classes \emph{not} present in the image (shown in gray) mislead the model into predicting a non-zero count. These spurious predictions for absent classes are often comparable to the true counts of present classes (shown in orange). 
    For instance, in the statistics produced using a SOTA model \cite{DBLP:journals/corr/abs-2407-04619}, the predicted number of \textit{cigarettes} exceeds that of \textit{blueberries}. (ii) On the right, we illustrate the \textit{distractor test}, where for each image the model is queried with each of its ground‑truth classes individually, i.e., the classes that are actually present in that image. This allows us to analyze how the model becomes distracted by the other co-occurring classes.}
    %\caption{\textbf{On the left}, the negative test aims at spotting the cases in which the classes non-present in the provided image (in gray) fool the model to output a non-zero number of instances. Often, these spurious counts from non-present classes are comparable to the number of instances of correct classes (in orange). For example, for a state-of-the-art model used to produce these statistics \cite{DBLP:journals/corr/abs-2407-04619}, there are more \textit{cigarettes} than \textit{blueberries}. \textbf{On the right}, for a given image, we probe the model with each of the ground-truth classes, one at a time, in order to understand \textit{how} the model is being \textit{distracted} by the other classes present within the same image.}
    \label{fig:teaser}
\end{figure*}

However, we identify major limitations in current benchmarks for evaluating text-guided CAC methods, which hinder both the accurate assessment of existing approaches and the development of more reliable future solutions. Our empirical analysis reveals that some state-of-the-art approaches often fail to determine \textit{which} object class should be counted based on the given prompt, highlighting a fundamental misalignment between the textual semantic description and the visual object features extracted by the model. This behavior emerges both when the prompt refers to an absent class---illustrated on the left side of Fig.~\ref{fig:teaser}---and when the image contains multiple object classes and the model is queried to count only one of them, as shown on the right side of Fig.~\ref{fig:teaser}.
This issue is particularly problematic in real-world applications. For instance, in an inventory management scenario, a system tasked with counting boxes of a specific product may instead count all visible packages, leading to inaccurate stock tracking. 

We attribute this shortcoming to two key factors: (i) deficiencies in current evaluation metrics and (ii) inherent limitations in existing CAC datasets. In fact, existing evaluation metrics, inherited from class-specific counting, focus solely on counting accuracy while neglecting a crucial aspect of text-guided CAC: the ability of the model to correctly understand the textual prompt. Additionally,  most CAC datasets consist predominantly of images containing a single object class, making it challenging to assess the ability of the model to distinguish between multiple object types within the same scene.

In this work, we address the two aforementioned limitations from complementary perspectives, thereby filling the existing gap. Specifically: (i) we propose a novel test suite which we name \benchmarkacron\ (\benchmarkname) to quantitatively assess the robustness and trustworthiness of text-guided CAC approaches; and (ii) we introduce \datasetacron\ (\datasetname), a new dataset for CAC that includes multiple object categories within each image, freely available to the scientific community~\cite{ciampi_2026_19231375}. \benchmarkacron\ goes beyond standard metrics by introducing two robustness tests tailored to probe failure modes under challenging conditions: (i) the \textit{negative-label} test (see Fig.~\ref{fig:teaser}, left), which assesses the ability of the model to reject misleading prompts by querying single-class images with references to absent categories; and (ii) the \textit{distractor} test (see Fig.~\ref{fig:teaser}, right), which evaluates the robustness of the model in multi-category scenarios where one object category serves as a distractor to the category described by the textual prompt.
%which evaluates artificially created mosaicked images built by stitching together pairs of single-class images, where one object category serves as a distractor to the category described by the textual prompt.
On the other hand, \datasetacron\ is a collection of 200 images containing common real-world objects, with the distinctive feature of including multiple object categories per image. This design overcomes the limitation of current CAC datasets, which include only single-category images, and provides the multi-category scenarios required for the distractor test described above.
%, thus overcoming the limitation of current datasets that contain only single-category images. 

To validate our contributions, we conduct an extensive experimental evaluation involving state-of-the-art open-world text-guided CAC techniques. Results reveal that several methods exhibit notable weaknesses in understanding object class descriptions, despite achieving top performance on standard class-specific metrics. Moreover, the presence of multiple object classes within the same image poses a significant challenge to many approaches.
Finally, we perform an additional analysis to investigate the reasons behind these errors. This includes not only qualitative examples but also a quantitative study of the relationship between the semantic similarity of the textual prompt and the performance of the evaluated approaches; the analysis reveals an occasionally measurable relationship between the two. We believe that our test suite and dataset will serve as a reference point for future research, highlighting the weaknesses of current models and emphasizing the need for more refined training procedures or even a reconsideration of their architectural designs.

Concretely, the contributions of this paper are as follows:
\begin{itemize}
    \item We empirically show that current evaluation protocols and datasets are insufficient for assessing the robustness of text-guided CAC methods, hindering the development of more effective solutions.
    \item We propose a new test suite, which we name \benchmarkacron\ (\benchmarkname), to evaluate the robustness and trustworthiness of existing open-world text-guided CAC models through two dedicated tests (negative-label and distractor tests), going beyond standard counting metrics. 
    \item We introduce and publicly release \datasetacron (\datasetname)~\cite{ciampi_2026_19231375}, a new collection of 200 real-world images featuring multiple object categories per image, thereby addressing the limitations of existing single-category datasets.
    \item We conduct an extensive experimental assessment involving SOTA open-world text-guided CAC approaches, showing a general and remarkable deficiency in understanding objects to be counted from the meaning of the textual prompt, despite achieving top results in standard class-specific counting metrics. We also analyze the difficulties these methods face in challenging multi-category scenarios and quantify how semantic similarity between the requested category and other categories correlates with performance.
\end{itemize}

This work extends our previous conference paper~\cite{DBLP:conf/wacv/CiampiMP0AF25}, which introduced an early‑stage test suite and preliminary results. We significantly expand upon that work in several directions by: (i) introducing \benchmarkacron, which revises and improves the initial PrACo test suite; (ii) presenting \datasetacron, a novel publicly available dataset featuring multiple object categories per image; (iii) conducting experiments on additional state‑of‑the‑art methods as well as on the new dataset; and (iv) providing an extensive analysis of the challenges these methods face in multi‑category scenarios, quantifying the relationship between semantic similarity among object categories and model performance.

We organize the remainder of this paper as follows. In Sec.~\ref{sec:related_works}, we review the most influential works on class-specific and class-agnostic counting, as well as existing CAC datasets and metrics. Sections~\ref{sec:test-suite} and~\ref{sec:dataset} present the main contributions of this paper---the test suite and the dataset. In Sec.~\ref{sec:experiments}, we report the experimental evaluation. Finally, Sec.~\ref{sec:conclusions} concludes the paper with insights into potential directions for future research.

%%%%%%%%%%%%%%%%%%%%%%%%%%%%%%%%%%%%%%%%%%%%%%%%%%%%%%%%%%%%%%%%%%%%%%%%%%%%%%%%%%%%%%%%%%%%%%
\section{Related Works}
\label{sec:related_works}

%\subsection{Class-specific Object Counting}
\subsection{Open-world Text-guided Class-agnostic Counting}
\label{sec:sec:related-work-sota}
Object counting is a fundamental task in computer vision, with broad applicability across various real-world domains. Consequently, numerous methods have been developed to count specific object categories, such as people~\cite{DBLP:conf/wacv/HossainHCW19,DBLP:journals/eswa/BenedettoCCFGA22,DBLP:conf/cvpr/LiuSF19,DBLP:journals/ivc/KhanMH23,DBLP:journals/ivc/ZhouH25,DBLP:journals/ivc/WangLQWZWW25,DBLP:journals/ivc/ZhouRLHS24}, vehicles~\cite{DBLP:conf/iccv/ZhangWCM17,DBLP:journals/eswa/CiampiGCFVA22,DBLP:conf/kst/SeenouvongWNKO16,8969620,DBLP:conf/ibpria/Guerrero-Gomez-Olmedo15}, insects~\cite{DBLP:journals/ecoi/CiampiZICBFAC23,BERECIARTUAPEREZ2022106933}, and biological cells~\cite{CIAMPI2022102500,8265200}. Approaches that regress and sum density maps have proven to be particularly effective in crowded scenarios~\cite{DBLP:conf/wacv/HossainHCW19,BERECIARTUAPEREZ2022106933,DBLP:conf/cvpr/LiuSF19,DBLP:conf/iccv/ZhangWCM17,8265200}, often outperforming methods based on direct object detection~\cite{DBLP:journals/eswa/CiampiGCFVA22,8969620}. Either way, the main disadvantage of class-specific object counting methods lies in their requirement for individually trained networks and, consequently, labeled datasets for each object type, limiting their applicability~\cite{DBLP:conf/cvpr/RanjanSNH21}.

%\subsection{Open-world Text-guided Class-agnostic Counting}
%\label{sec:sec:related-work-sota}
Thus, recent research in object counting has increasingly focused on class-agnostic approaches, tackling the above-mentioned limitations while aiming to reduce annotation effort and generalize across arbitrary object categories~\cite{DBLP:conf/cvpr/RanjanSNH21}. Among the different possibilities to specify the target object category to count, the open-world text-guided paradigm expresses it as a textual description~\cite{DBLP:conf/aaai/0018C24,DBLP:conf/aaai/KangMKH24,10483595,DBLP:conf/mm/JiangLC23,DBLP:conf/cvpr/XuL0RS23,AminiNaieni23}, offering greater flexibility and emerging as a key direction for future progress, despite its current performance gap compared to the exemplar-based paradigm, where the category is expressed as visual exemplars~\cite{DBLP:conf/bmvc/LiuZZX22,DBLP:journals/corr/abs-2405-11770,DBLP:conf/aaai/WangX0024,Dukic_2023_ICCV,DBLP:journals/tip/WuCLCWL25,DBLP:journals/spl/GongYZ25,DBLP:journals/ivc/XuLYWLZ25,DBLP:journals/ivc/ZhangZCWH25}.

Several open-world text-guided CAC methods rely on vision–language models that align text and image features in a shared space, then decode them into density maps. For example,~\cite{DBLP:conf/cvpr/XuL0RS23} adds a conditional VAE on CLIP~\cite{DBLP:conf/icml/RadfordKHRGASAM21} to generate exemplar prototypes from category semantics. CounTX~\cite{AminiNaieni23} predicts counts directly using patch–text similarity, while CLIP-Count~\cite{DBLP:conf/mm/JiangLC23} propagates semantics across multiple resolutions via hierarchical patch–text interactions. VLCounter~\cite{DBLP:conf/aaai/KangMKH24} fine-tunes CLIP with modules that exploit intermediate features.
More recently, DAVE~\cite{Pelhan_2024_CVPR} proposed a two-stage detect-and-verify paradigm: the first stage generates a high-recall set of candidate detections, while the second stage refines these predictions through unsupervised clustering and CLIP-based verification. In contrast to the above architectures, TFPOC~\cite{10483595} adopts a detection-driven strategy leveraging SAM~\cite{kirillov2023segment} for instance segmentation. Its pipeline has two steps: (i) an enhanced CLIP variant generates visual exemplar prototypes via image–text similarity; (ii) similarity maps are computed between image features and SAM masks, prompted by bounding boxes from the first stage. 
Similarly, PseCo~\cite{DBLP:conf/cvpr/HuangD0ZS24} employs a detection-based pipeline with SAM for segmentation, but instead of uniform grid prompts, it uses class-agnostic localization to generate a heatmap from which object coordinates guide SAM. A second stage then leverages CLIP to classify segmented regions based on exemplars.
More, GroundingREC~\cite{10656642} introduces Referring Expression Counting (REC), enabling fine-grained distinctions within the same category. Their method adapts the open-set detector GroundingDINO~\cite{DBLP:conf/eccv/LiuZRLZYJLYSZZ24} using the CLS token to represent global semantics of the referring expression.
%, replacing bounding boxes with box centers and using the CLS token to represent global semantics of the referring expression. 
CountGD~\cite{DBLP:journals/corr/abs-2407-04619} introduces a single-stage open-world counting model that accepts text prompts, visual exemplars, or both. Built on GroundingDINO, it extends the architecture with modules that fuse exemplar and text tokens via self- and cross-attention. Finally, UPC~\cite{DBLP:conf/aaai/0018C24} unifies boxes, points, and text into a prompt mask, applies cross-attention for density estimation, and iteratively refines predictions using a fixed-point loss and contrastive training for robustness.

However, most of these open-world text-guided CAC methods largely stem from exemplar-based approaches, where the target category is indicated by selecting exemplars from the image. Consequently, these methods assume the object is present in the scene. In this work, we challenge this assumption by exploring scenarios where (i) the object may be absent, (ii) multiple object classes may coexist, or (iii) the query may be ambiguous or misleading. Our experiments reveal that current open-world text-guided models still respond to non-present categories and struggle in multi-class scenarios, underscoring their limited ability to accurately interpret object categories from textual descriptions.

\subsection{Existing Datasets and Metrics}
\label{sec:sec:related-work-dataset}
There is a lack of publicly available datasets for CAC. Most existing datasets focus on specific categories, such as UCF-QNRF~\cite{DBLP:conf/eccv/IdreesTAZARS18} for crowd counting; NDISPark~\cite{DBLP:conf/visapp/CiampiSCGA21, ciampi_ndispark_6560823} for vehicles; Pest Sticky Traps~\cite{ciampi_2023_7801239} for pest monitoring; and VGG Cell~\cite{DBLP:journals/cmbbeiv/XieNZ18} for cell estimation. Conversely, multi-category datasets like MSCOCO~\cite{DBLP:conf/eccv/LinMBHPRDZ14} are unsuitable for counting, as they were designed for detection and typically contain only a few object instances per image.

Among the few datasets available for CAC, the gold standard is FSC-147~\cite{DBLP:conf/cvpr/RanjanSNH21}, which contains 6,135 images across 147 categories (plants, animals, vehicles, food). Each image typically contains objects from a single category. Objects are annotated with dot-based centroids, and each image includes three exemplar bounding boxes always belonging to a single category and a text file with the category name. A revised version of FSC-147, called FSC-133, was introduced in~\cite{DBLP:journals/corr/abs-2205-10203}, correcting several errors and providing 5,898 images across 133 categories. Despite these improvements, most works still use FSC-147 as the standard. Another variant, FSC-147-D~\cite{AminiNaieni23}, replaces simple object descriptions provided by class names with more fine-grained and structured natural language sentences. Similarly, REC-8K~\cite{DBLP:conf/cvpr/DaiLC24} enriches textual descriptions with attributes across about 8,000 images drawn from existing datasets, including FSC-147. The only exceptions addressing the limitation of single-category images are OmniCount-191~\cite{DBLP:conf/aaai/MondalNZ025} and MCAC~\cite{DBLP:conf/eccv/HobleyP24}, which introduce multi-category scenarios. However, OmniCount‑191 contains few objects per image and is largely derived from video sequences, resulting in highly similar consecutive frames and many visually similar object instances, while MCAC is synthetic, not realistic, and lacks the annotations required for open‑world text‑guided approaches. Still within the synthetic domain,~\cite{DBLP:conf/eccv/DAlessandroMH24} and~\cite{DBLP:conf/wacv/DoubinskyACB24} leverage text-to-image latent diffusion models to generate counting data across a diverse range of object categories.

Counting performance is typically assessed using mean absolute error (MAE) and root mean squared error (RMSE), which capture absolute and squared errors, respectively: $\text{MAE} = \frac{1}{N} \sum_{n=1}^{N} \left| \tilde{c}n - c_n \right|$ and \\ $\text{RMSE} = \sqrt{\frac{1}{N} \sum{n=1}^{N} ( \tilde{c}_n - c_n )^2}$, where $N$ is the number of test images, and $\tilde{c}n$ and $c_n$ denote ground truth and predicted counts. While MAE reflects average error magnitude, RMSE penalizes larger errors more heavily. 
%A less common alternative is MAPE, a normalized variant of MAE, defined as $\text{MAPE} = \frac{1}{N} \sum{n=1}^{N} \frac{|\tilde{c}_n - c_n|}{\tilde{c}_n}$. 
Although these metrics are standard, they ignore spatial distribution, so models can achieve low errors while misplacing objects. To address this, the Grid Average Mean Absolute Error (GAME)~\cite{DBLP:conf/ibpria/Guerrero-Gomez-Olmedo15} is adopted. GAME divides the image into $4^L$ regions (grid level $L$) and sums the MAE per region, thus considering both count and coarse localization.

However, current datasets and evaluation protocols pose major limitations for benchmarking open-world text-guided CAC methods. To address this, we 
%build upon our previous work~\cite{DBLP:conf/wacv/CiampiMP0AF25} and 
introduce \datasetacron, a new dataset featuring multiple object categories per image, along with \benchmarkacron, a test suite that goes beyond traditional metrics inherited from class-specific counting. Unlike these conventional evaluators, which suffer from severe shortcomings, our suite is designed to assess the robustness and trustworthiness of existing open-world text-guided CAC models.

%%%%%%%%%%%%%%%%%%%%%%%%%%%%%%%%%%%%%%%%%%%%%%%%%%%%%%%%%%%%%%%%%%%%%%%%%%%%%%%%%%%%%%%%%%%%%%
\section{The \benchmarkacron\ (\benchmarkname) Test Suite}
\label{sec:test-suite}

\begin{figure*}[!htbp]
    \centering
    \includegraphics[width=0.95\linewidth]{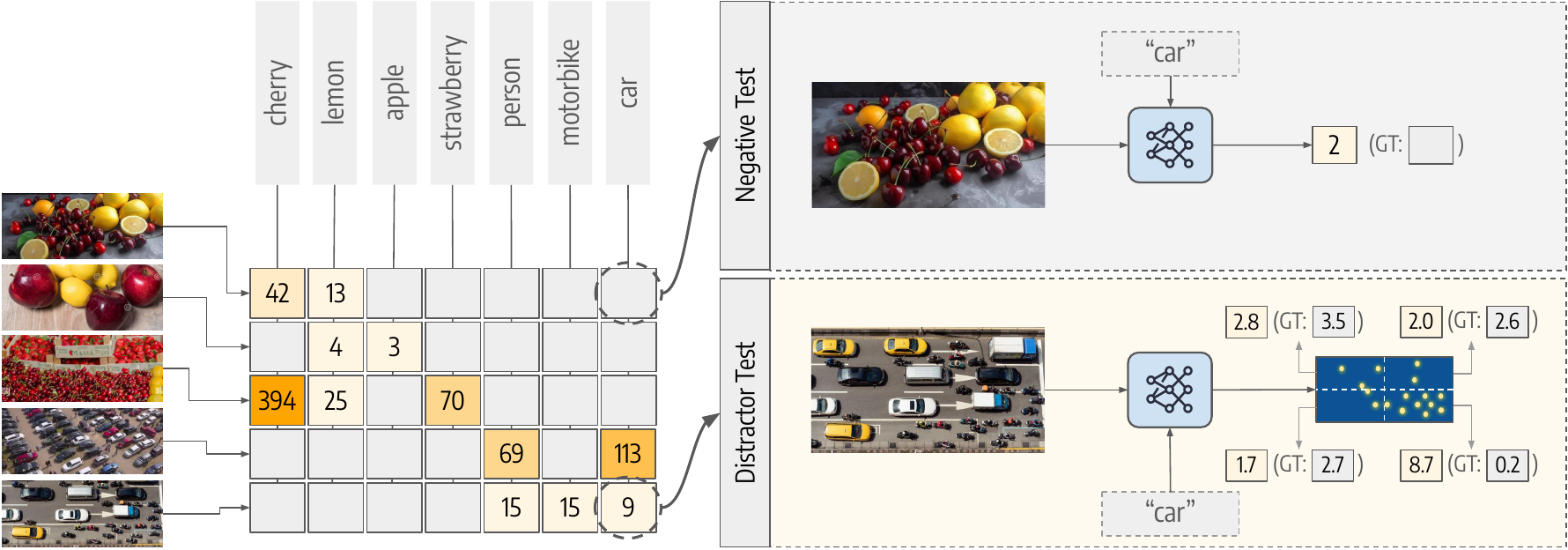}
    \caption{\textbf{Overview of the inference procedure used in the test suites.}
    For each image, the model predicts object counts for all categories in the dataset. Orange boxes report the ground-truth counts for categories present in the image, while gray boxes denote categories not present (ground-truth count equal to zero). In the \textit{negative-label} test, we consider only categories not present in the image (gray boxes) and evaluate how much the predicted count deviates from zero. In the \textit{distractor} test, we focus on categories present in the image and analyze where counting errors occur by computing dedicated counting metrics over spatial subregions.}
    \label{fig:method}
\end{figure*}

\subsection{Overview and Formal Framework} 
The main goal of our proposed test suite is to move beyond the narrow evaluation protocols commonly used for text‑guided CAC---which typically focus solely on the counting error of a specific object category present in an image---and instead probe the semantic grounding and trustworthiness of the models. In other words, we aim to determine whether open‑world text‑guided CAC systems truly \textit{understand} the user‑provided prompts describing target object classes, or whether they effectively function as high‑level saliency detectors that count the most visually prominent objects in a scene, regardless of their actual categories.
%To address the multi-category nature of our newly introduced dataset, we extend the formal evaluation framework to account for images containing multiple distinct object classes. 

To formalize our framework, let $\mathcal{I} = \{I_1, I_2, \dots, I_N\}$ represent a collection of $N$ images. We further consider a text-guided CAC model $\mathcal{M}(I, t)$ that, given an image $I$ and a textual description $t$, produces a predicted count $c$. The textual description, or \textit{prompt}, is directly derived from the target object class name through simple templating, e.g., \textit{``A [class] in the image''}, as commonly adopted in text-guided CAC approaches.

For every image $I_i \in \mathcal{I}$, we define two distinct sets of prompts:
\begin{itemize}
    \item $\mathcal{P}_i = \{p_{i,1}, p_{i,2}, \dots, p_{i,K_i}\}$: This represents the set of \textit{positive prompts}, each of which contain one among the $K_i$ object categories that are actually present within the image $I_i$.
    \item $\mathcal{N}_i = \{n_{i,1}, n_{i,2}, \dots, n_{i,M_i}\}$: This represents the set of \textit{negative prompts}, which consist of all the dataset categories that are entirely absent from the scene. Therefore, $\mathcal{N}_i = \mathcal{C} \setminus \mathcal{P}_i$, where $\mathcal{C}$ is the set of prompts defined for the specific dataset.
\end{itemize}

In an ideal, semantically robust system, the output of the model should satisfy two primary conditions:
\begin{enumerate}
    \item For any positive prompt $p \in \mathcal{P}_i$, the output $\mathcal{M}(I_i, p)$ should approximate the ground-truth count $\tilde{c}(I_i, p)$ for that specific class.
    \item For any negative prompt $n \in \mathcal{N}_i$, the output $\mathcal{M}(I_i, n)$ should ideally be $0$, indicating that the model correctly recognizes the absence of that category.
    \item For any per-image positive prompt $p \in \mathcal{P}_i$, the output $\mathcal{M}(I_i, p)$ should not interfere with the count on the other instances from different classes within the same image $\{\mathcal{M}(I_i, p')\}$, where $p' \in \mathcal{P}_i \setminus p$ is a \textit{distractor} for the image $i$ when prompted with $p$.
\end{enumerate}

Existing benchmarks focus almost exclusively on the first condition, effectively ignoring how a model behaves when asked to count objects that are not present in an image. Our suite enables systematic evaluation on both single‑class and multi‑class CAC datasets, allowing us to extensively test the remaining two conditions and provide a more complete and precise characterization of the shortcomings of different text-guided CAC models across datasets.
In particular, we introduce two complementary tests. The \textit{negative‑label} test evaluates condition~\textit{2}, i.e., the ability of text-guided CAC models to correctly ignore object categories that are absent from an image. Its goal is to identify cases in which models fail to properly ground the textual prompt and instead count objects indiscriminately. Conversely, the \textit{distractor} test evaluates condition~\textit{3}, assessing the resilience of a text-guided model in accurately counting the requested class when other instances from distractor categories are present in the same image. Thus, this test provides a more nuanced analysis of model failures in multi‑category scenarios. We graphically show these two tests in Fig.~\ref{fig:method}, and we formally develop them in the following sections.

%In particular, we introduce two complementary tests. The \textbf{Negative-label test} aims at evaluating condition 2, which is the capability of CAC models to ignore any non-present class within an image; its aim is to detect whenever models are failing in understanding the textual prompt or counting every object indiscriminately. Conversely, the \textbf{Distractor test} aims at evaluating condition 3., which is the resilience of the model in counting the requested class when other instances from distractor classes are present within the same image, introducing a more nuanced control over the model failure modes. We present these two proposed tests in detail in the following sections.
%two targeted tests which evaluate, from different perspectives, the second condition. In particular, the \textbf{Negative-label test} aims at evaluating the capabilities of CAC models to ignore any non-present class within an image. Conversely, the \textbf{Distractor test} aims at evaluating the resilience of the model in counting the requested class when other instances from distractor classes are present within the same image.

\subsection{Negative-Label Test}
The \textit{negative-label} test is designed to quantify the ability of the model to ignore objects not related to the input prompt. In practical applications---such as inventory management or autonomous surveillance---it is just as vital for a model to correctly report a zero count for absent items as it is to provide accurate counts for present ones. A model that yields a high count for \textit{``cats''} when presented with an image containing only \textit{``cars''} is not semantically grounded, even if its \textit{``cars''} count is highly accurate. Such a model is likely biased toward counting any salient object it encounters, irrespective of the textual input. To quantitatively estimate this aspect, we perform exhaustive cross-probing across all images with all non-present (negative) classes. Specifically, for every image $I_i$ in the dataset, we query the model with every prompt defined in the negative set $\mathcal{N}_i$ to check how much the model activates with object classes surely not present in the image. 
While the open-world nature of the evaluated models allows them to process any arbitrary textual prompt, we purposely restrict the set of negative queries to the categories $C$ defined within the dataset. As a result, it avoids ambiguities arising from unannotated background objects that could otherwise be mistakenly penalized as model hallucinations, thereby preserving the integrity and reproducibility of the benchmark.

To provide a standardized quantitative assessment of this behavior, we propose the following two specialized metrics.

\subsubsection{Normalized Mean of Negative predictions (NMN)}
The \textit{NMN} provides an aggregate measure of a model's tendency to over-count when prompted with non-existent classes. To ensure this metric is comparable across images with varying levels of density and crowdedness, we normalize the erroneous \textit{false} counts by the total number of actual objects present in the scene. This normalization is based on the empirical observation that models often scale their error relative to the visual complexity of the image; counting five non-existent objects in a scene containing hundreds of real ones is a different degree of failure than counting five non-existent objects in a scene that is nearly empty. Formally, let $T_i = \sum_{p \in \mathcal{P}_i} \tilde{c}(I_i, p)$ be the total sum of ground-truth object instances across all positive categories in image $I_i$. The NMN for a dataset of $N$ images is defined as follows:

\begin{equation}
\text{NMN} = \frac{1}{N} \sum_{i=1}^{N} \left( \frac{1}{|\mathcal{N}_i|} \sum_{n \in \mathcal{N}_i} \frac{\mathcal{M}(I_i, n)}{T_i} \right).
\end{equation}

In this formulation, $|\mathcal{N}_i|$ represents the number of negative prompts evaluated for image $I_i$. A lower NMN value indicates a higher degree of textual sensitivity and a lower propensity for hallucinating counts based on visual saliency alone.

\subsubsection{Positive Class Count Nearness (PCCN)}
While NMN captures the average counting error, it may obscure catastrophic failure cases in which a model exhibits severe semantic misalignment—producing a non-zero count for a negative class that is numerically closer to the total number of objects in the image than the count predicted for the correct positive class.
%While NMN offers a view of the average error, it may mask individual catastrophic failures where a model is completely "lost"—meaning it produces a count for a negative class that is more "convincing" (closer to the actual number of total objects in the image) than its count for the correct positive class. 
\textit{PCCN} measures the frequency of these semantic breakdowns. We define two distance metrics for each image $I_i$ to facilitate this comparison:
\begin{enumerate}
    \item $d^{\text{pos}}_i$: The average absolute error for all positive prompts present in the image
    \begin{equation}
    d^{\text{pos}}_i = \frac{1}{|\mathcal{P}_i|} \sum_{p \in \mathcal{P}_i} | \mathcal{M}(I_i, p) - \tilde{c}(I_i, p) |.
    \end{equation}
    
    \item $d^{\text{neg}}_i$: For each positive class, the mean distance to all negative predictions, then averaged across positive classes 
    \begin{equation}
        d^{\text{neg}}_i = \frac{1}{|\mathcal{P}_i|} \sum_{p \in \mathcal{P}_i} \left[ \frac{1}{|\mathcal{N}_i|} \sum_{n \in \mathcal{N}_i} |\mathcal{M}(I_i, n) - \tilde{c}(I_i, p)| \right].
    \end{equation}    
\end{enumerate}

The PCCN is then calculated as the percentage of images in the dataset where the prediction of the model for the correct classes is closer to the ground truth than its ``erroneous'' predictions for absent classes. %are to the total salient object count:
\begin{equation}
    \text{PCCN} = \frac{1}{N} \sum_{i=1}^{N} \mathbb{I}(d^{\text{pos}}_i < d^{\text{neg}}_i) \cdot 100\%.
\end{equation}

Here, $\mathbb{I}(\cdot)$ is the indicator function, which equals $1$ if the condition is true and $0$ otherwise. While high PCCN percentages (approaching $100\%$) are not sufficient to declare that the model can ignore negative classes, what's more interesting is the case where models exhibit low PCCN. In fact, a low PCCN indicates critical failures, with, on average, a higher count from negative classes than from positive ones. This may indicate a model that is heavily misunderstanding or even ignoring the given textual prompt, resulting in a highly unreliable output.

\subsection{Distractor Test}
\label{sec:distractor-test}
The \textit{distractor test} is designed to evaluate the robustness of text‑guided class‑agnostic counting (CAC) models when identifying a target object category in the presence of other, potentially confusing object instances (i.e., distractors). While the negative‑label test assesses the ability to reject prompts referring to absent categories, the distractor test focuses on the discriminative capability of a model in heterogeneous scenes. This setting reflects realistic scenarios, where objects rarely appear in isolation and models must selectively attend to the queried category (e.g., ``\textit{apples}'') while ignoring co‑occurring distractors (e.g., ``\textit{oranges}'' or ``\textit{bottles}'').

\subsubsection{Multi-class and Mosaic Approaches}
We consider two complementary implementations of the distractor test, depending on the nature of the available data.

\paragraph{Direct Multi-class Evaluation}
Using a multi-class counting dataset (such as the one introduced in Section~\ref{sec:dataset}), models are evaluated on real‑world images containing multiple object categories. Given an image $I_i$ and a target prompt $p \in \mathcal{P}_i$, all remaining categories $p' \in \mathcal{P}_i \setminus \{p\}$ naturally act as distractors.

\paragraph{Mosaic-Based Evaluation}
To leverage existing single\-class CAC datasets, we adopt a synthetic augmentation strategy inspired by~\cite{AminiNaieni23}. Specifically, we construct mosaicked images $I_{ij}^{\text{mosaic}} = \text{vstack}(I_i, I_j)$ by vertically concatenating a \emph{positive} image $I_i$ containing the target class and a \emph{negative} image $I_j$ containing a different class. This setup induces a controlled spatial partitioning, in which the upper region contains all true positives, while the lower region contains only distractors. Although this approximation introduces a single negative class per image, it provides a simple and effective mechanism to evaluate confusion in standard single‑class CAC benchmarks.

For both settings, we assume that the model $\mathcal{M}$ produces a density map or spatially grounded predictions for a given prompt. For density‑based methods, the predicted count over a region $\Omega$ is obtained by integrating the density map, i.e., $c_\Omega = \int_{\Omega} \mathcal{D}(x,y)\, dA$. These spatial predictions enable the computation of distraction-specific metrics, detailed in the following, which require incorporating some spatial awareness rather than relying solely on the final count.

\subsubsection{Counting Precision and Recall}
To quantitatively evaluate performance in complex multi-class scenes under the \textit{distractor} test, it is crucial to have, at least approximately, access to the spatial origin of the predicted counts for each class. Relying solely on final per-class count outputs, without any information about \textit{where} instances are detected, may be insufficient to assess the robustness of the model to class confusion. For instance, consider an image containing three \textit{pears} and three \textit{apples}: a model may correctly output a count of three for both classes, while internally misassigning \textit{pears} as \textit{apples} and vice versa. For these reasons, we introduce \textit{counting precision} and \textit{counting recall} metrics.

Specifically, we drew inspiration from precision and recall metrics commonly used in \textit{detection} scenarios, where we can precisely assess the correctness of each predicted instance. However, a significant divergence exists between prompt-based CAC and standard object detection frameworks. In detection, every discrete proposal can be explicitly validated as correct or incorrect via spatial overlap with bounding-box annotations. Conversely, in CAC, we are not expected to generate precise instance-level localizations, as the objective is to estimate an aggregated global count for a specific textual category. This necessitates a strategic adaptation of the concepts of true positives ($TP$), false positives ($FP$), and false negatives ($FN$) to operate over regional density estimates or patch-based counts rather than individual localized detections.

\begin{figure*}[!htbp]
    \centering
    \includegraphics[width=0.95\columnwidth]{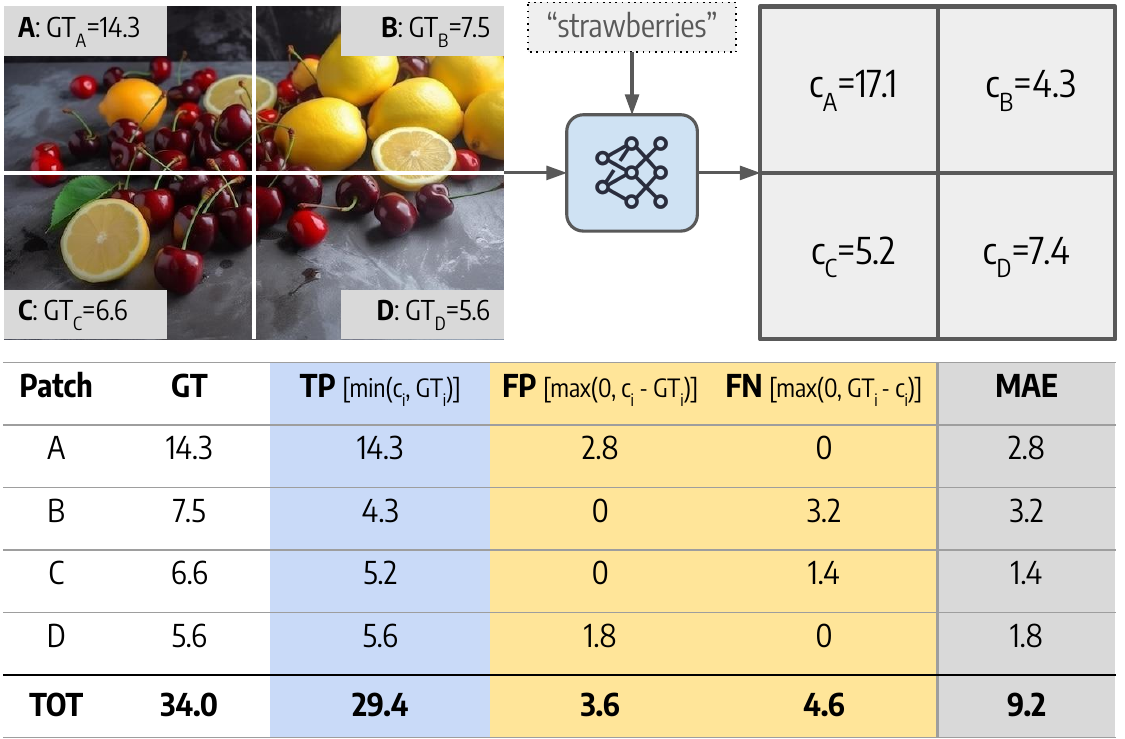}
    \caption{\textbf{Example illustrating the computation of the proposed counting metrics on the \textit{distractor} test.} 
    The image is partitioned into spatial patches, and the output density map is integrated within each patch to obtain patch-level predicted counts. For each patch, we compute true positives (TP), false positives (FP), false negatives (FN), and the mean absolute error (MAE), and subsequently aggregate their contributions. The patch-wise TP, FP, and FN values are summed to compute image-level counting precision, recall, and F1-score. In contrast, summing the MAE over all patches yields the well-known GAME metric, which we also report in the experimental results. Note that, unlike detection-based metrics, counting precision and recall naturally take continuous (non-integer) values.}
    \label{fig:counting-example}
\end{figure*}
%\clearpage

Following the spatial partitioning strategy of~\cite{DBLP:conf/ibpria/Guerrero-Gomez-Olmedo15} for the computation of the GAME metric (see also Sec.~\ref{sec:related_works}), each image is divided into a grid of $W = 4^L$ non‑overlapping patches $\mathcal{G} = \{g_1, \dots, g_W\}$, where $L \in \mathbb{N}$ denotes the grid level. By default, we use $L = 1$, while results for finer granularities are reported in~\ref{sec:additional-results}.

Then, for a given patch $g_w$ in image $I_i$ when prompted with class $p_i$, let $c_{i,w}$ be the predicted count and $\tilde{c}_{i,w}$ be the ground-truth count. We define the patch-wise components as follows:
\begin{itemize}
    \item $\text{TP}_{i,w} = \min(c_{i,w}, \tilde{c}_{i,w})$: The portion of the predicted count that correctly matches the ground-truth instances of the target class within the patch.
    %The portion of the prediction that matches the actual presence of the target class.
    \item $\text{FP}_{i,w} = \max(0, c_{i,w} - \tilde{c}_{i,w})$: The over-estimation of the predicted count, corresponding to false-positive instances within the patch, i.e., the number of distractor objects within the patch.
   % The over-estimation in a patch, representing counts of distractor objects.
    \item $\text{FN}_{i,w} = \max(0, \tilde{c}_{i,w} - c_{i,w})$: The under-estimation of the predicted count, corresponding to false-negative instances within the patch, i.e., the number of target objects within the patch that were not detected.
    %The under-estimation, representing instances of the target class that the model failed to detect.
\end{itemize}

To obtain a robust indicator of dataset-wide performance, we aggregate these localized values across all spatial patches and across the test images. In particular, for each image $I_i$, we accumulate true positives (TP), false positives (FP), and false negatives (FN) over the $W$ image patches as follows:
\begin{equation}
\begin{aligned}
\text{TP}_i = \sum_{w=1}^W \text{TP}_{i,w} \quad \text{FP}_i= \sum_{w=1}^W \text{FP}_{i,w} \quad \text{FN}_i = \sum_{w=1}^W \text{FN}_{i,w}
\end{aligned}
\end{equation}

Finally, we derive the \textit{Counting Precision (CntP)}, \textit{Counting Recall (CntR)}, and the \textit{Counting F1-score (CntF1)} over the entire dataset averaging image-level scores:
%, ensuring that each test image contributes equally to the final evaluation:
\begin{equation}
\text{CntP} = \frac{1}{N} \sum_{i=1}^{N}\frac{\text{TP}_i}{\text{TP}_i + \text{FP}_i}
\end{equation}
\begin{equation}
\text{CntR} = \frac{1}{N} \sum_{i=1}^{N}\frac{\text{TP}_i}{\text{TP}_i + \text{FN}_i}
\end{equation}
\begin{equation} 
\text{CntF1} = \frac{1}{N} \sum_{i=1}^{N}\frac{2 \cdot \text{TP}_i}{2 \cdot \text{TP}_i + \text{FP}_i + \text{FN}_i} 
\end{equation}

We report an example showcasing the computation of these quantities in Fig.~\ref{fig:counting-example}. Intuitively, the \textit{CntR} measures the sensitivity of the model: a low recall indicates that the model struggles to identify target instances, often due to the presence of distracting visual patterns that divert the attention of the model from the target class. The \textit{CntP} reflects the precision of the model: a low precision indicates that the model incorrectly includes distractor categories in the predicted count for a given prompt. Finally, the \textit{CntF1} represents the harmonic mean of \textit{CntP} and \textit{CntR}, providing a single summary metric of the ability of the model to operate reliably in complex, multi-class environments. 

Note that the proposed metrics extend those introduced in our previous work~\cite{DBLP:conf/wacv/CiampiMP0AF25}, which were formulated within a mosaic-based evaluation protocol and limited to single-class datasets. By removing the constraint of a fixed two‑patch structure, the proposed definition naturally extends to arbitrary spatial partitions, while preserving the same underlying assumptions. We formally show in Sec.~\ref{sec:conting-prec-rec} that the original mosaic‑based precision–recall formulation emerges as a special case of the proposed framework.

%%%%%%%%%%%%%%%%%%%%%%%%%%%%%%%%%%%%%%%%%%%%%%%%%%%%%%%%%%%%%%%%%%%%%%%%%%%%%%%%%%%%%%%%%%%%%%
\section{The \datasetacron\ (\datasetname) Dataset}
\label{sec:dataset}

\begin{figure*}[!htbp]
    \centering
    \begin{subfigure}{0.24\textwidth}
        \centering
        \begin{tcolorbox}[imgbox, colframe=red]
            \includegraphics[width=\linewidth,height=3cm]{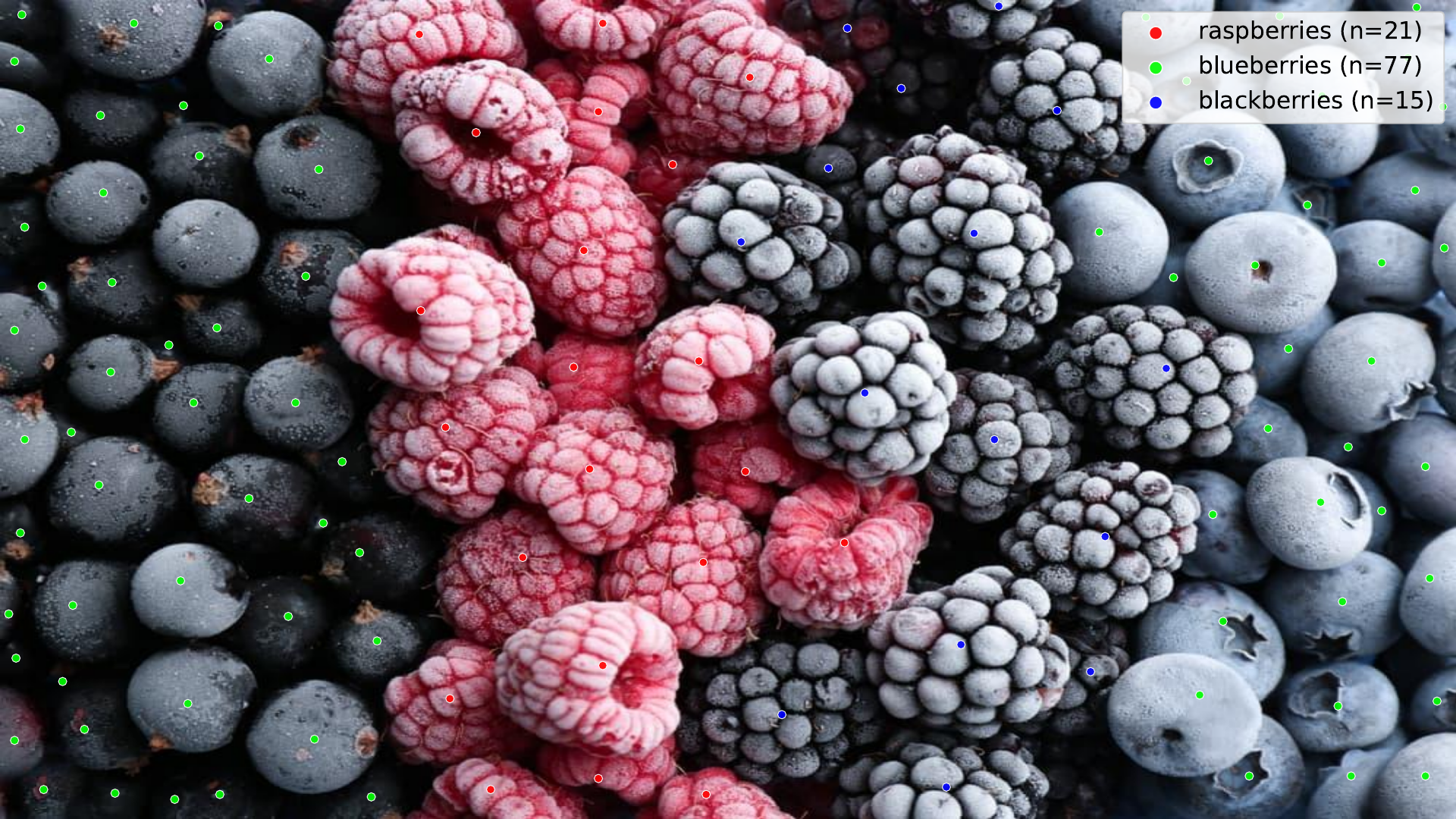}
        \end{tcolorbox}
    \end{subfigure}
    \begin{subfigure}{0.24\textwidth}
        \centering
        \begin{tcolorbox}[imgbox, colframe=blue]
            \includegraphics[width=\linewidth,height=3cm]{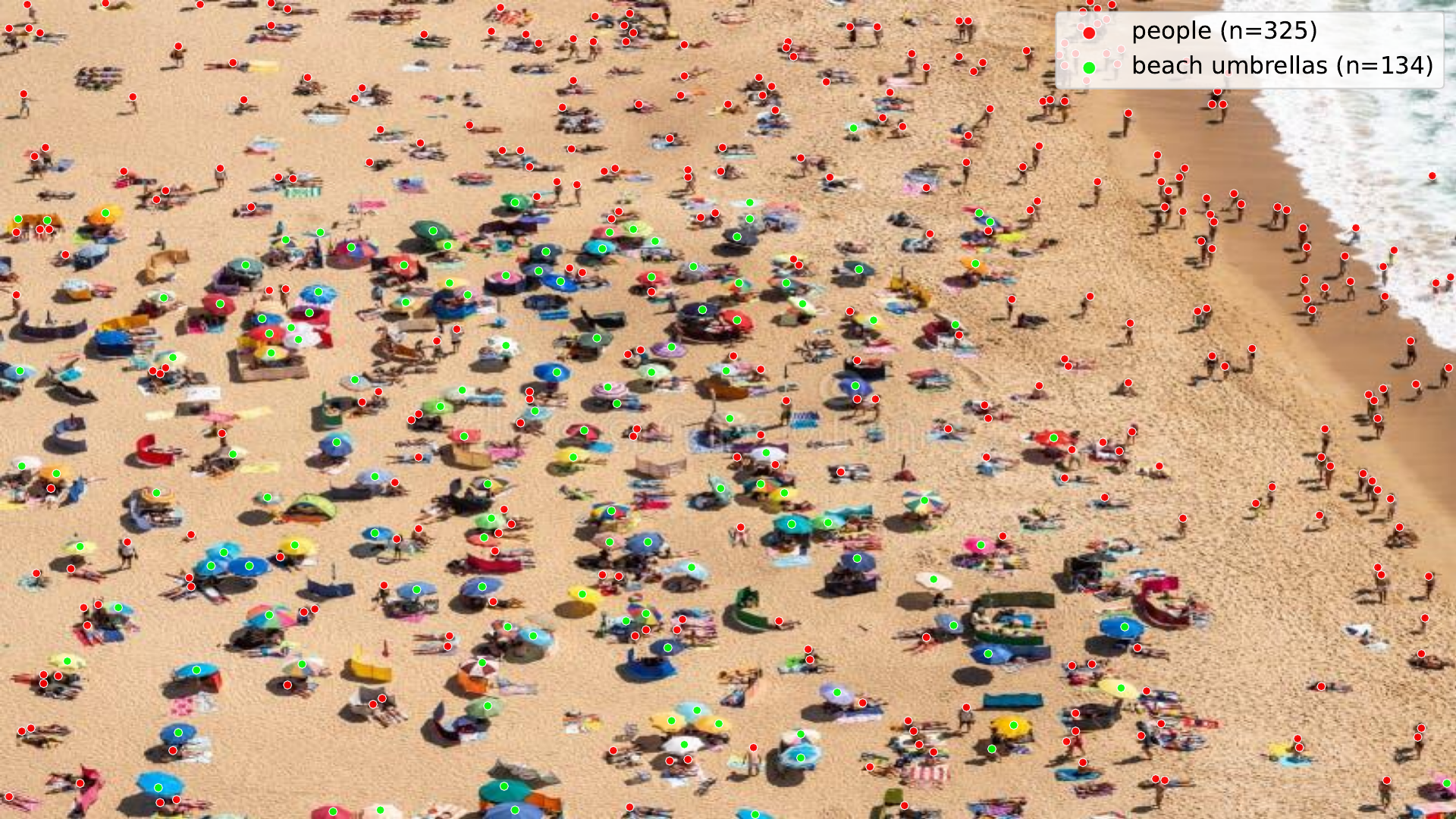}
        \end{tcolorbox}
    \end{subfigure}
    \begin{subfigure}{0.24\textwidth}
        \centering
        \begin{tcolorbox}[imgbox, colframe=cyan]
            \includegraphics[width=\linewidth,height=3cm]{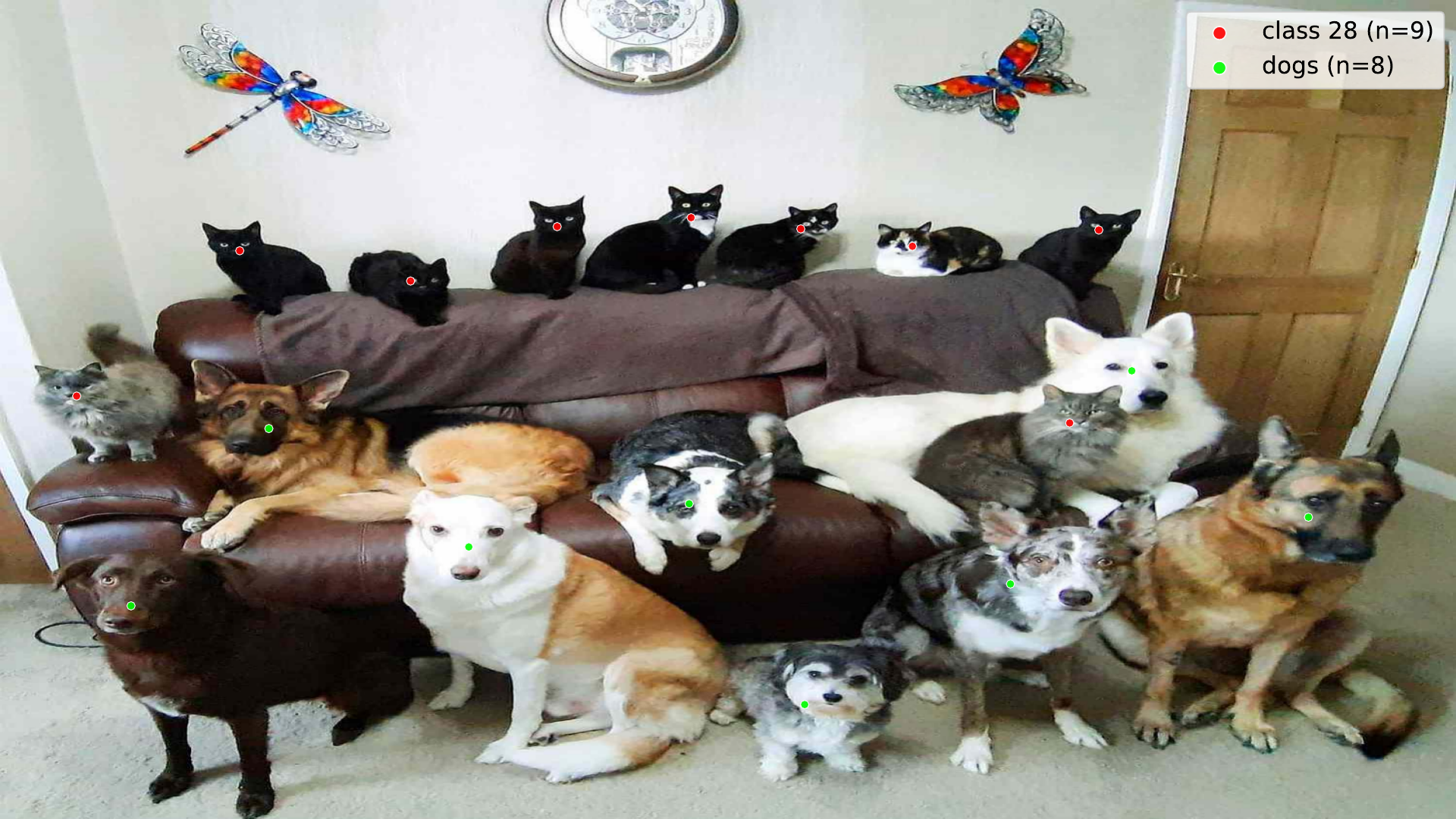}
        \end{tcolorbox}
    \end{subfigure}
    \begin{subfigure}{0.24\textwidth}
        \centering
        \begin{tcolorbox}[imgbox, colframe=green]
            \includegraphics[width=\linewidth,height=3cm]{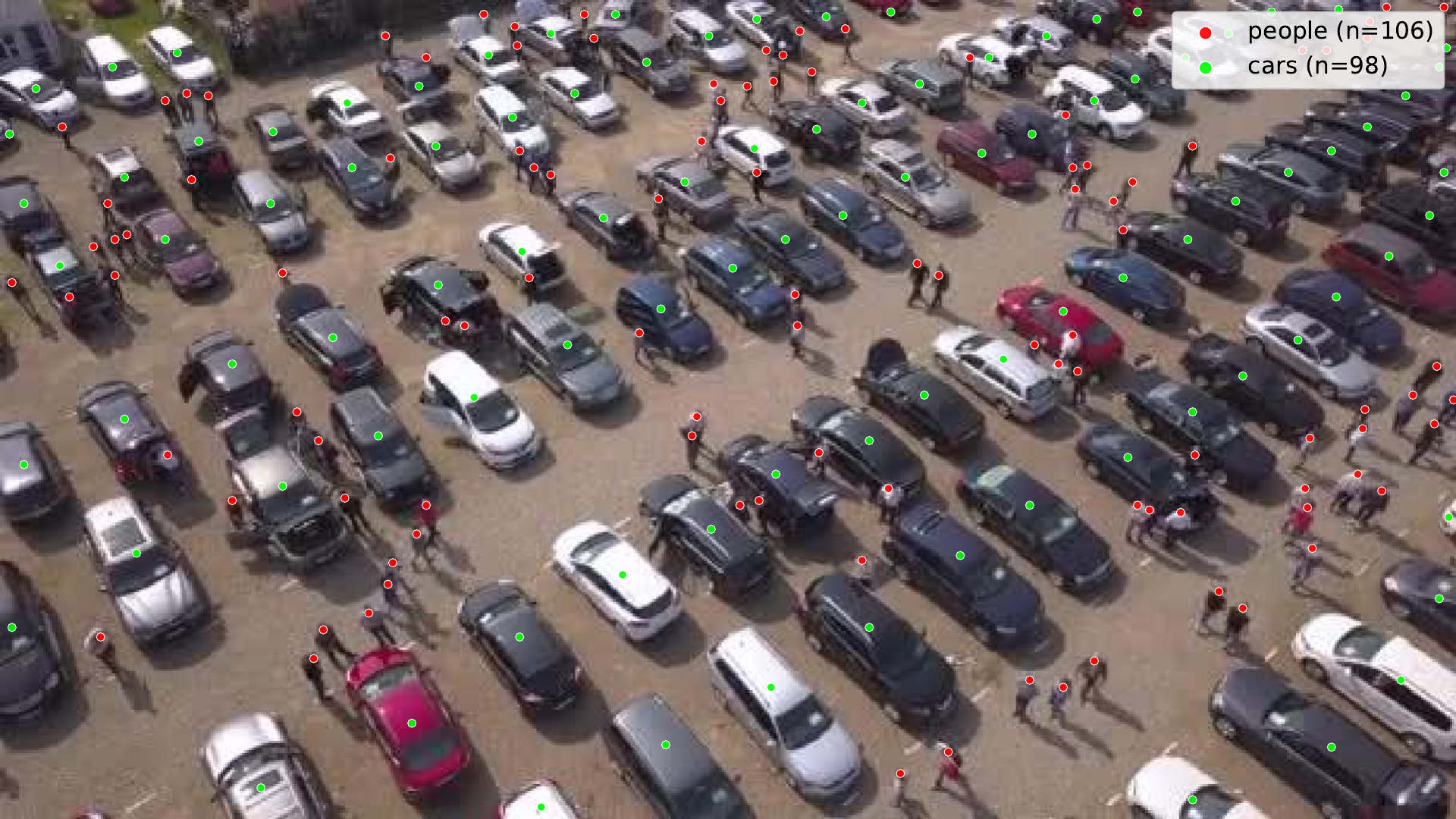}
        \end{tcolorbox}
    \end{subfigure}
    \vspace{0.1cm}  % aumenta/diminuisci a piacere
    \begin{subfigure}{0.24\textwidth}
        \centering
        \begin{tcolorbox}[imgbox, colframe=purple]
            \includegraphics[width=\linewidth,height=3cm]{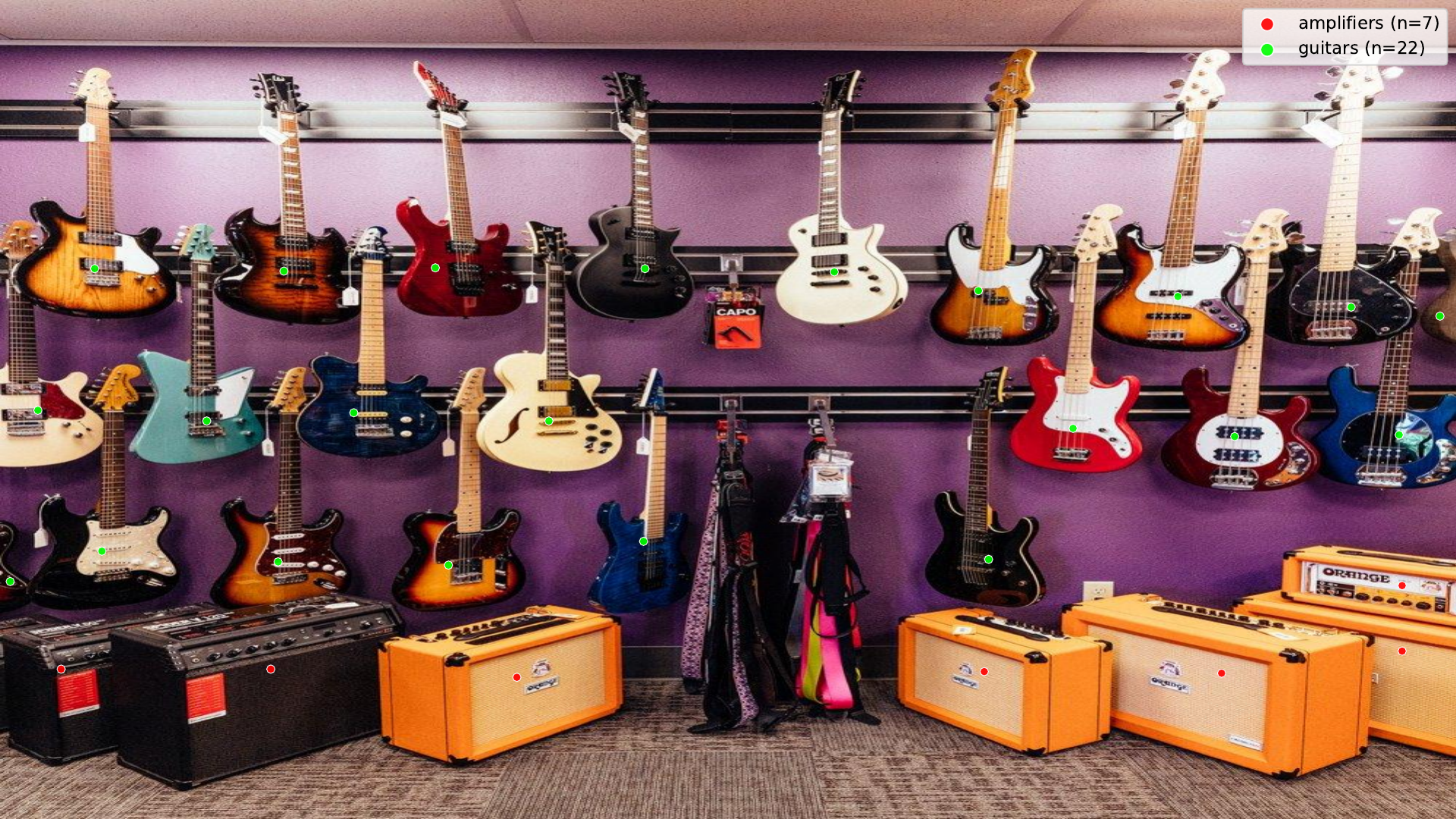}
        \end{tcolorbox}
    \end{subfigure}
    \begin{subfigure}{0.24\textwidth}
        \centering
        \begin{tcolorbox}[imgbox, colframe=brown]
            \includegraphics[width=\linewidth,height=3cm]{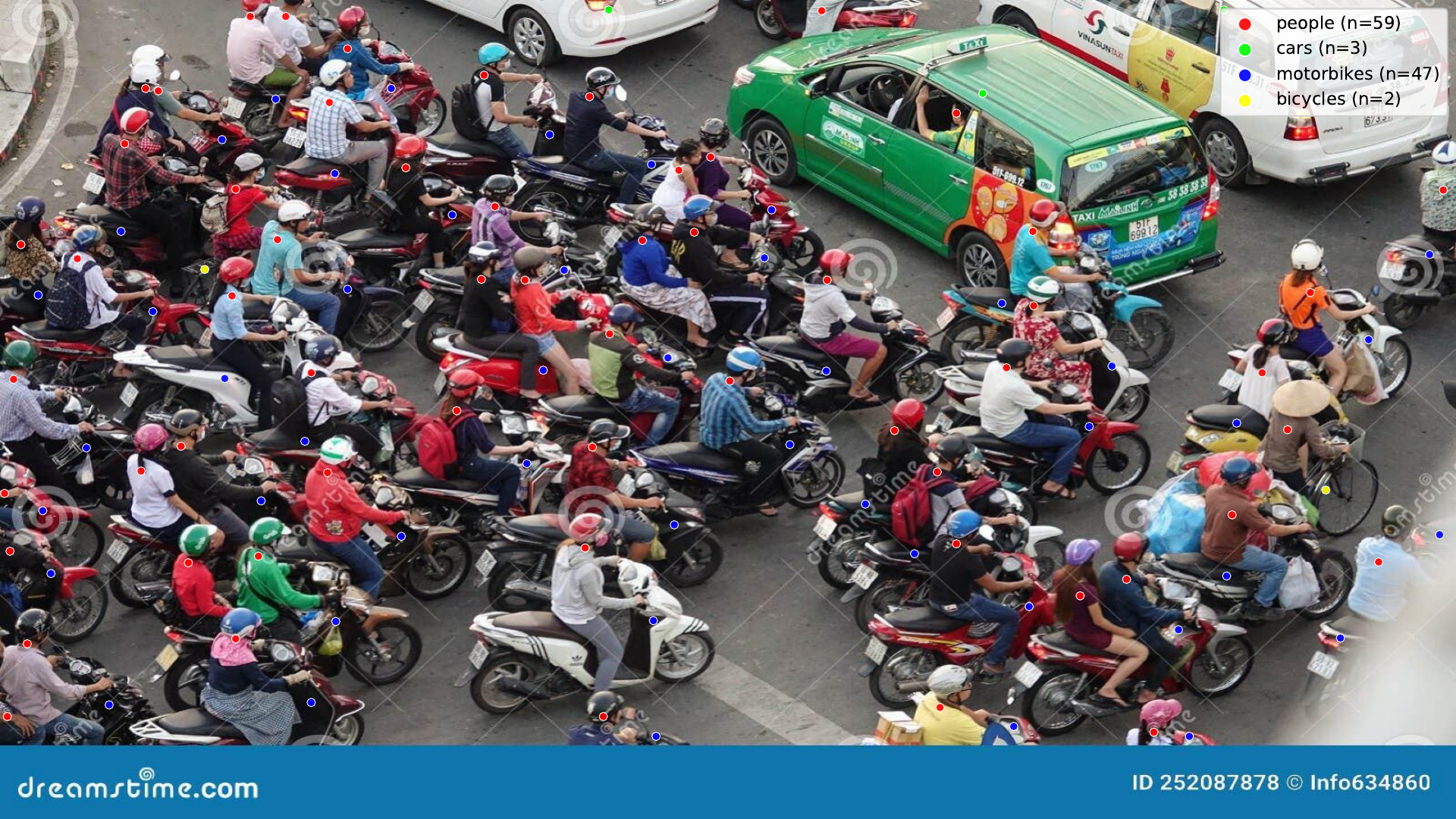}
        \end{tcolorbox}
    \end{subfigure}
    \begin{subfigure}{0.24\textwidth}
        \centering
        \begin{tcolorbox}[imgbox, colframe=orange]
            \includegraphics[width=\linewidth,height=3cm]{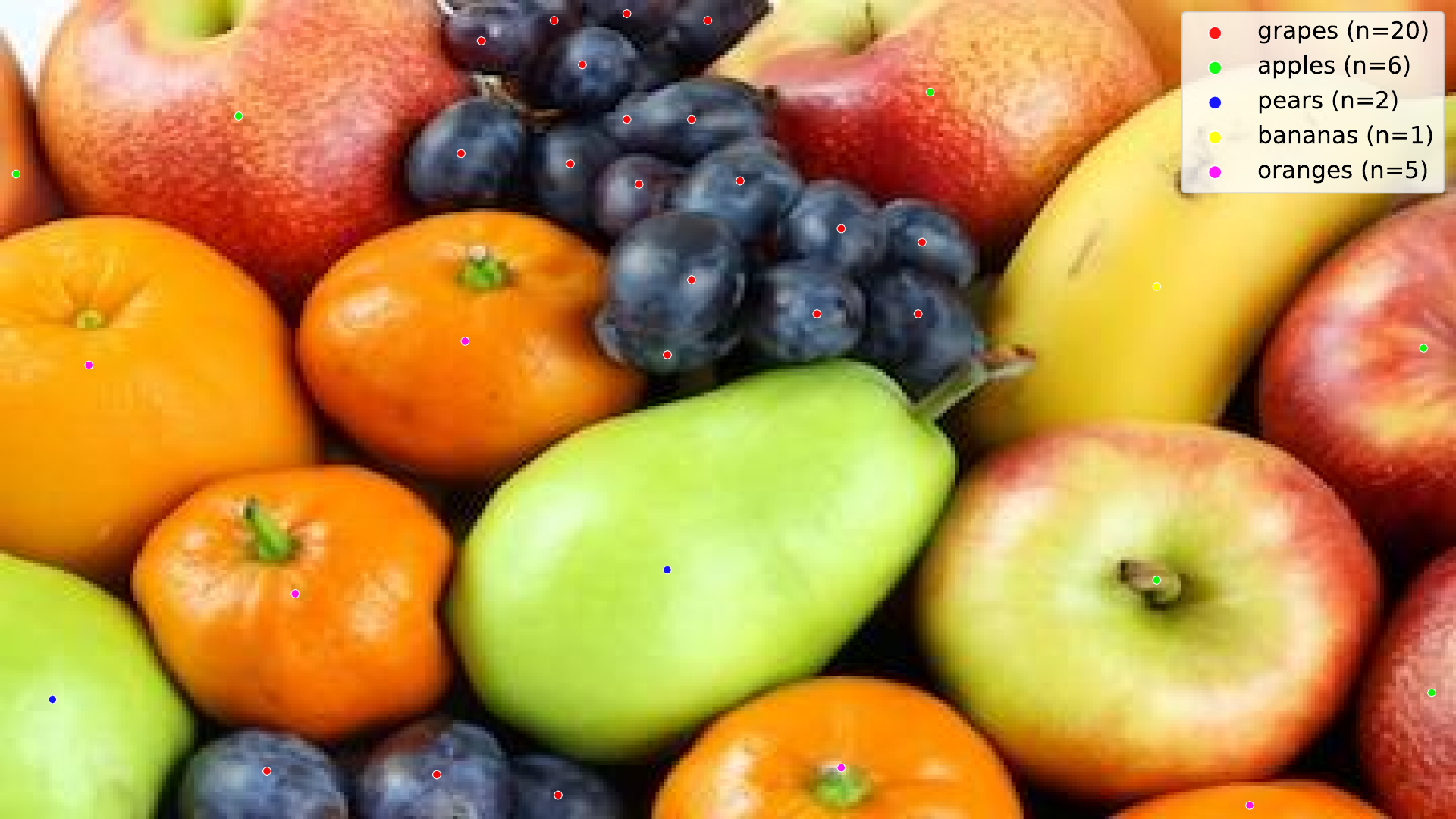}
        \end{tcolorbox}
    \end{subfigure}
    \begin{subfigure}{0.24\textwidth}
        \centering
        \begin{tcolorbox}[imgbox, colframe=magenta]
            \includegraphics[width=\linewidth,height=3cm]{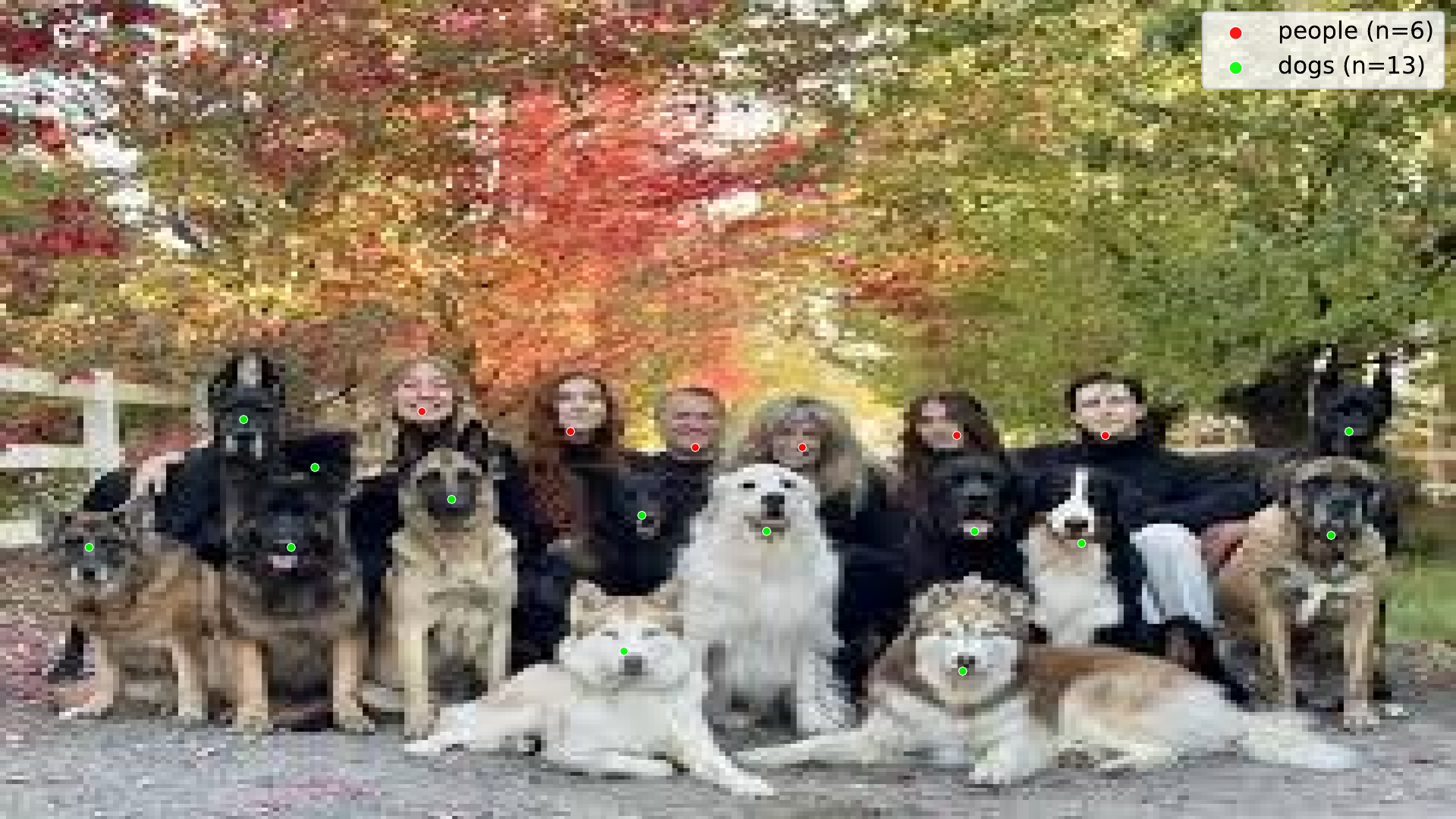}
        \end{tcolorbox}
    \end{subfigure}
    \caption{\textbf{Samples from the \datasetacron{} dataset.} 
    We show a selection of dot-annotated images from our \datasetname{} (\datasetacron) dataset, a collection specifically designed for CAC and characterized by the presence of multiple object categories annotated within each image.}
    \label{fig:mucca_samples}
\end{figure*}

\begin{figure*}[!t]
    \centering
    \includegraphics[width=\linewidth]{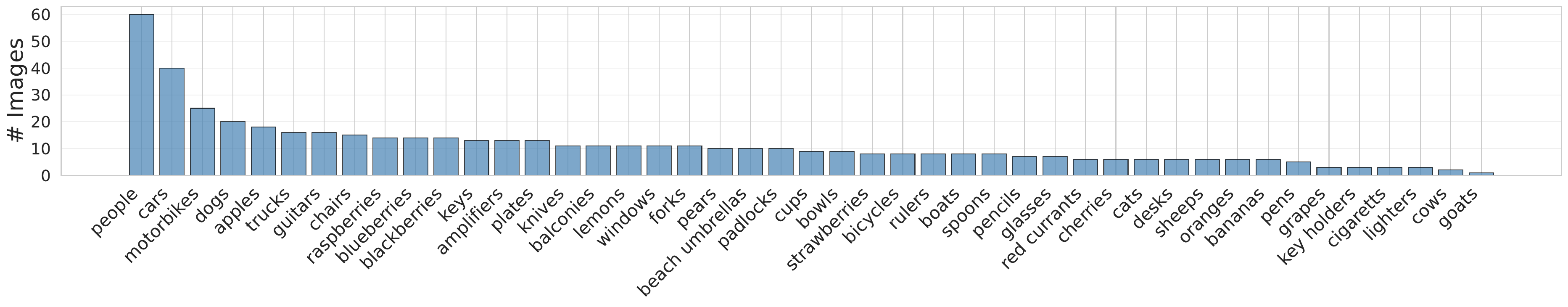}
    \includegraphics[width=0.45\linewidth]{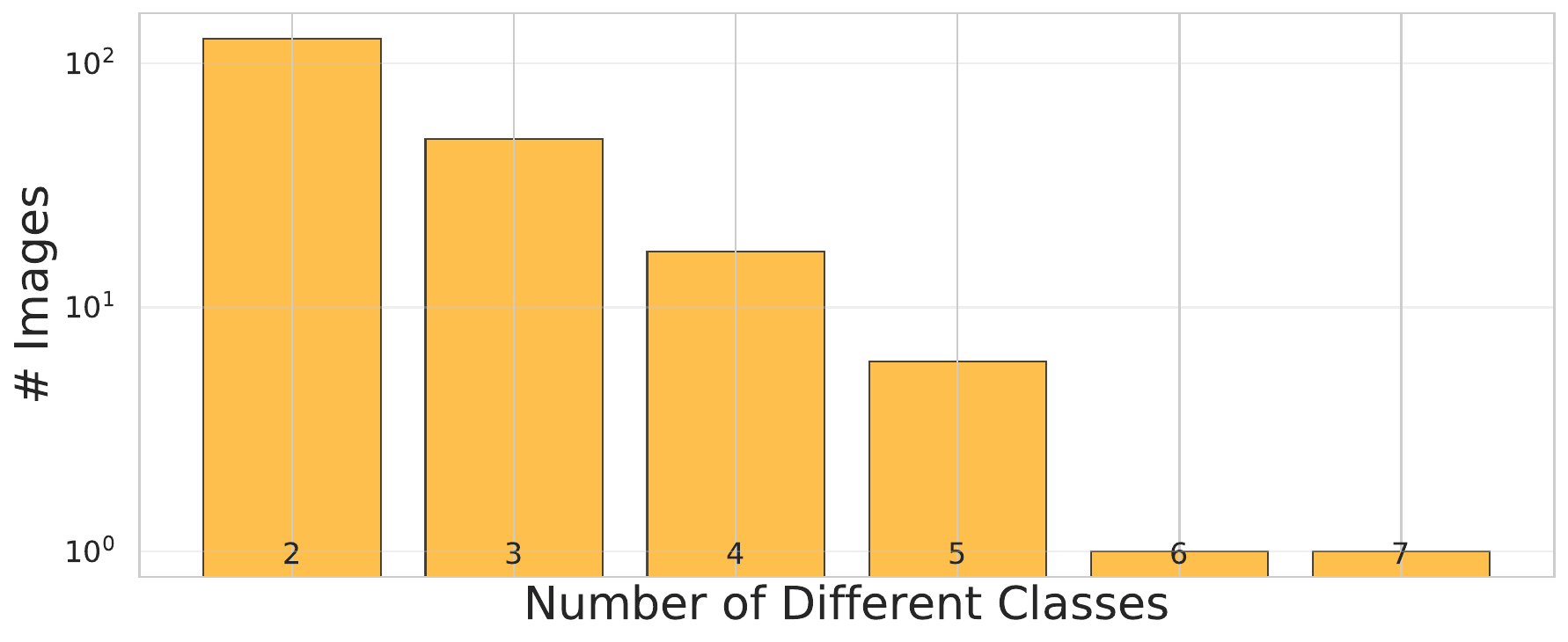}
    \hspace{0.05\linewidth}
    \includegraphics[width=0.45\linewidth]{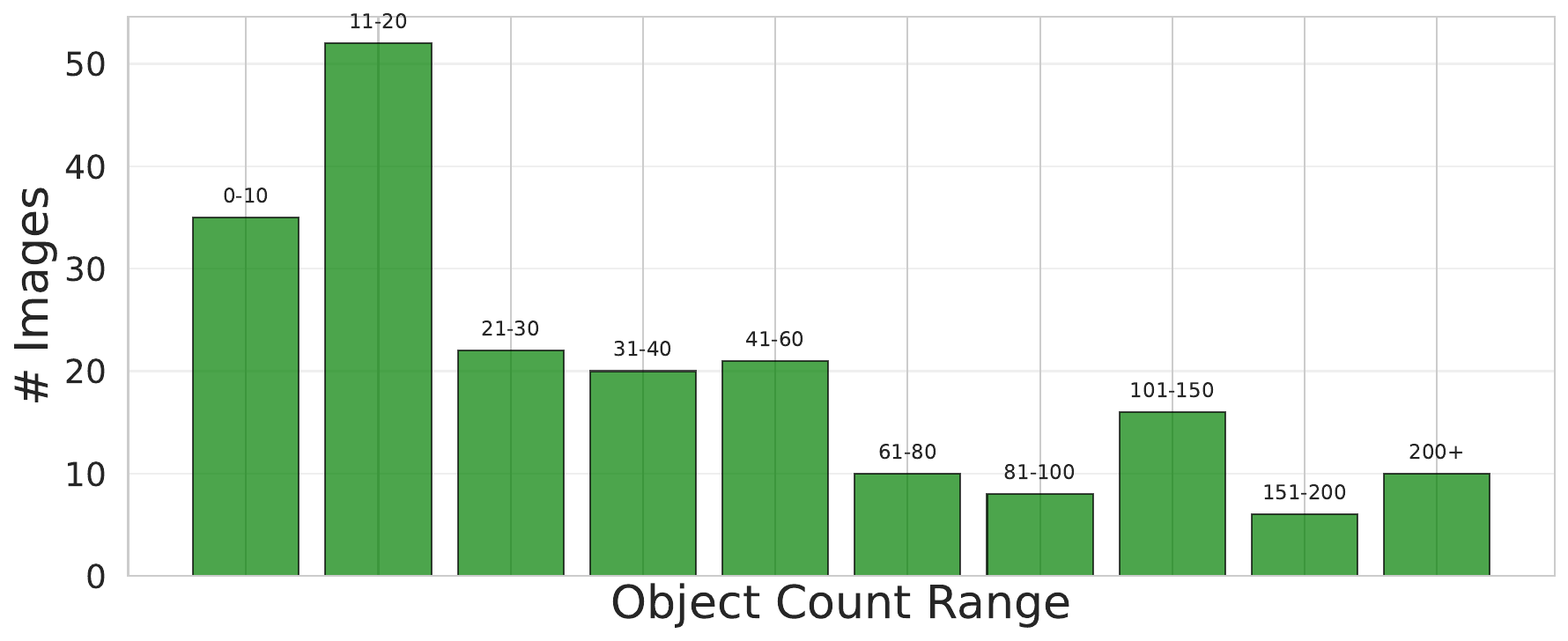}
    \caption{\textbf{Statistical overview of the MUCCA dataset.}
    Top: frequency of object classes measured by the number of images in which each class appears.
    Bottom-left: distribution of images according to the number of distinct object categories they contain, highlighting the intrinsic multi-class nature of the dataset.
    Bottom-right: distribution of the total number of object instances per image, illustrating the wide variability in scene density across the dataset.}
    \label{fig:dataset-stats}
\end{figure*}

Public datasets for CAC remain extremely limited. The gold standard, FSC‑147~\cite{DBLP:conf/cvpr/RanjanSNH21}, contains images that typically include objects from a single category, and the annotations themselves are restricted to one class per image. We argue that this design choice indeed represents a fundamental limitation, as it prevents a meaningful evaluation of CAC methods in scenarios that require distinguishing multiple object types within the same scene. Only very recently have multi‑category datasets been introduced, such as OmniCount‑191~\cite{DBLP:conf/aaai/MondalNZ025} and MCAC~\cite{DBLP:conf/eccv/HobleyP24}. However, these datasets suffer from significant limitations: they are either synthetic---thus lacking full realism---or derived from video sequences, resulting in highly similar consecutive frames and many visually similar object instances. To fill this gap, we collect and publicly release \datasetacron\ (\datasetname)~\cite{ciampi_2026_19231375}, a new dataset specifically designed for open-world text-guided CAC that provides images with annotations for multiple object categories within each scene. 

\paragraph{Data collection and curation} Images were collected from the web using Google search queries involving multiple object categories that reflect realistic contextual and semantic dependencies, e.g., \textit{``images of markets with many berries''}. The collected images were then filtered to remove those containing additional object categories not intended for counting (particularly in the background), as well as images of insufficient quality. Subsequently, each image was manually annotated by placing a dot at the most representative point of every object instance belonging to the target classes, following annotation practices commonly adopted in class-specific and CAC benchmarks~\cite{DBLP:conf/nips/LempitskyZ10,DBLP:conf/cvpr/RanjanSNH21}. Dot annotations were associated with specific object categories, resulting in category‑specific sets of dots. The annotation process was carried out by two annotators, who cross‑checked each other’s work to ensure annotation accuracy. In addition to dot annotations, we also provide the corresponding class names, which can be directly used as textual queries for open‑world text‑guided CAC methods. Some annotated samples are illustrated in Fig.~\ref{fig:mucca_samples}.

\paragraph{Dataset statistics} 
The dataset comprises 200 images and 11,576 annotated instances spanning 45 distinct object classes. The images exhibit substantial variability in object density, with an average of 57.88 objects per image and a median of 26.5. Overall, the dataset spans a broad spectrum of difficulty levels, ranging from sparse scenes containing as few as 3 objects to highly crowded environments with up to 955 instances. A key characteristic of the benchmark is its intrinsic multi‑class nature: each image contains between 2 and 7 distinct object categories. This diversity is further enriched by natural semantic co‑occurrences; for example, ``cars'' and ``people'' frequently appear together (13\% of images), while ``blueberries'' and ``raspberries'' co‑occur in fruit collections (6.5\%), reflecting realistic contextual dependencies.
The 45 object classes follow a long‑tail distribution typical of natural scenes. The class ``people'' is the most prevalent, appearing in 30\% of the images with a total of 3,756 instances, often forming dense crowds. In contrast, categories such as ``bicycles'' and ``cows'' are much more sparsely represented. High‑density classes like ``cherries'' (with an average of 95.5 instances per image) and ``beach umbrellas'' pose particularly challenging scenarios for small‑object counting, whereas larger objects such as ``cars'' and ``chairs'' provide medium‑density counting challenges. A visual summary of these statistics is provided in Fig.~\ref{fig:dataset-stats}.

%%%%%%%%%%%%%%%%%%%%%%%%%%%%%%%%%%%%%%%%%%%%%%%%%%%%%%%%%%%%%%%%%%%%%%%%%%%%%%%%%%%%%%%%%%%%%%
\section{Experimental Evaluation}
\label{sec:experiments}

\subsection{Experimental Setting}
\label{sec:sec:experiments-setting}
We conduct two sets of experiments using our \benchmarkacron\ test suite on two datasets, covering both the negative‑label test and the distractor test. For the first set of experiments, we employ the FSC‑147 dataset~\cite{DBLP:conf/cvpr/RanjanSNH21}, the gold standard for CAC, which contains 6,135 images spanning 147 categories (see Sec.~\ref{sec:sec:related-work-dataset} for further details). Since FSC‑147 is a single‑class dataset---i.e., images typically contain objects belonging to a single category, and annotations refer to only one category per image---we adopt the \textit{Mosaic} implementation of the distractor test, described in Sec.~\ref{sec:distractor-test}. Although most images contain objects of a single category, we filter out the few multi‑class images (for which, in any case, annotations are not provided) by following the procedure in~\cite{Pelhan_2024_CVPR}. This prevents interference with our proposed tests, ensuring that no false positives arise from additional object classes present in the same image.
For the second set of experiments, we instead use our new \datasetacron{} dataset, which is natively multi‑class and therefore enables direct multi‑class evaluation for the distractor test. 

\begin{table*}[!t]
    \centering
    \caption{\textbf{Results on the \textit{test set} of the FSC-147 dataset~\cite{DBLP:conf/cvpr/RanjanSNH21}.} We report the performance of the ten SOTA methods analyzed in this work using our \benchmarkacron{} test suite (negative-label and distractor tests), along with standard counting-specific evaluation metrics. Best, second-best, and third-best results are indicated with \gold, \silver, and \bronze, and formatted using \textbf{bold}, \textit{italic}, and \underline{underline}, respectively.}
    \vspace{3pt}
    \label{tab:results-FSC147-test-set}%
    \newcolumntype{L}{>{\arraybackslash}m{.27\linewidth}}%
    \newcolumntype{C}{>{\centering\arraybackslash}X}
    %\tiny%
    \setlength{\tabcolsep}{0.8pt}
    \begin{tabularx}{0.98\linewidth}{L|CC|CCC|CCC}
        \toprule
        & \multicolumn{2}{c|}{Negative-label Test} & \multicolumn{3}{c|}{Distractor Test} & \multicolumn{3}{c}{Classic} \\
        \cmidrule(lr){2-3} \cmidrule(lr){4-6} \cmidrule(lr){7-9}
        \centering Method & {\footnotesize NMN} $\downarrow$ & {\footnotesize PCCN $\uparrow$} &
        {\footnotesize CntP $\uparrow$} & {\footnotesize CntR $\uparrow$} &
        {\footnotesize CntF1 $\uparrow$} & {\footnotesize GAME(1) $\downarrow$} &
        {\footnotesize MAE $\downarrow$} & {\footnotesize RMSE $\downarrow$} \\
        \midrule
        \midrule
        ZSC~\cite{DBLP:conf/cvpr/XuL0RS23} {\footnotesize (CVPR '23)} & 1.04 & 48.82 & 0.50 & \underline{0.82}\bronze & 0.57 & 65.20 & 21.41 & 131.83 \\
        %CounTX~\cite{AminiNaieni23} {\footnotesize (BMVC '23)} & 0.96 & 64.45 & 0.66 & 0.72 & 0.63 & 51.13 & 15.93 & 106.28 \\
        CounTX~\cite{AminiNaieni23} {\footnotesize (BMVC '23)} & 0.95 & 64.51 & 0.66 & 0.72 & 0.63 & 51.13 & \underline{15.92}\bronze & \underline{106.89}\bronze\mbox{} \\
        %CLIP-Count~\cite{DBLP:conf/mm/JiangLC23} {\footnotesize (ACM MM '23)} & 1.28 & 37.98 & 0.49 & 0.75 & 0.55 & 66.07 & 17.61 & 109.33 \\
        CLIP-Count~\cite{DBLP:conf/mm/JiangLC23} {\footnotesize (ACM MM '23)} & 1.27 & 38.13 & 0.49 & 0.75 & 0.55 & 66.07 & 17.59 & 109.97 \\
        %VLCounter~\cite{DBLP:conf/aaai/KangMKH24} {\footnotesize (AAAI '24)} & 1.17 & 53.11 & 0.50 & 0.78 & 0.57 & 64.14 & 17.10 & 106.43 \\
        VLCounter~\cite{DBLP:conf/aaai/KangMKH24} {\footnotesize (AAAI '24)} & 1.15 & 53.36 & 0.50 & 0.78 & 0.57 & 64.14 & 17.02 & 106.93 \\
        %TFPOC~\cite{10483595} {\footnotesize (WACV '24)} & 0.76 & 65.55 & 0.68 & 0.83 & 0.69 & 45.35 & 24.65 & 137.26 \\
        TFPOC~\cite{10483595} {\footnotesize (WACV '24)} & 0.75 & 66.04 & 0.68 & 0.83 & 0.69 & 45.35 & 24.79 & 138.11 \\
        %DAVE~\cite{Pelhan_2024_CVPR} {\footnotesize (CVPR '24)} & \textbf{0.08} & \textbf{97.23} & 0.78 & 0.73 & 0.72 & 36.59 & \textbf{15.28} & \textbf{104.31} \\
        DAVE~\cite{Pelhan_2024_CVPR} {\footnotesize (CVPR '24)} & \textbf{0.08}\gold & \textbf{97.62}\gold & \underline{0.78}\bronze & 0.73 & \underline{0.72}\bronze & \underline{36.59}\bronze & \textbf{15.23}\gold & \textbf{103.53}\gold\mbox{} \\
        %\rowcolor{yellow}
        %PseCo~\cite{DBLP:conf/cvpr/HuangD0ZS24} {\footnotesize (CVPR '24)} & 1.05 & 53.11 & 0.33 & 0.48 & 0.35 & 96.50 & 17.11 & 133.07 \\
        PseCo~\cite{DBLP:conf/cvpr/HuangD0ZS24} {\footnotesize (CVPR '24)} & 1.05 & 53.11 & 0.53 & \textit{0.85}\silver & 0.61 & 61.07 & 17.08 & 133.05 \\
        %GroundingREC~\cite{10656642} {\footnotesize (CVPR '24)}$^*$ & 0.37 & 93.28 & \textbf{0.86} & \underline{0.85} & \textbf{0.84} & \textbf{28.00} & 20.46 & 134.48 \\
        GroundingREC~\cite{10656642} {\footnotesize (CVPR '24)} & \underline{0.35}\bronze & \underline{92.86}\bronze & \textbf{0.89}\gold & 0.80 & \textbf{0.82}\gold & \textbf{28.86}\gold & 22.67 & 135.79 \\
        UPC~\cite{DBLP:conf/aaai/0018C24} {\footnotesize (AAAI '24)} & 1.21 & 61.43 & \textit{0.80}\silver & \underline{0.82}\bronze & \textit{0.78}\silver & \textit{33.81}\silver & \textit{15.82}\silver & \textit{104.40}\silver\mbox{} \\
        CountGD~\cite{DBLP:journals/corr/abs-2407-04619} {\footnotesize (NeurIPS '24)} & \textit{0.11}\silver & \textit{93.36}\silver & 0.74 & \textbf{0.86}\gold & \textit{0.78}\silver & 38.23 & 15.95 & 132.09 \\
        \bottomrule 
    \end{tabularx}

    %* usando il loro dataset grande; $\$$ usando il loro fsc
\end{table*}

\begin{table*}[!t]
    \centering
    \caption{\textbf{Results on the \textit{validation set} of FSC-147 dataset~\cite{DBLP:conf/cvpr/RanjanSNH21}.} We report the performance of the ten SOTA methods analyzed in this work using our \benchmarkacron{} test suite (negative-label and distractor tests), along with standard counting-specific evaluation metrics. Best, second-best, and third-best results are indicated with \gold, \silver, and \bronze, and formatted using \textbf{bold}, \textit{italic}, and \underline{underline}, respectively.}
    \vspace{3pt}
    \label{tab:results-FSC147-val-set}%
    \newcolumntype{L}{>{\arraybackslash}m{.27\linewidth}}%
    \newcolumntype{C}{>{\centering\arraybackslash}X}
    %\tiny%
    \setlength{\tabcolsep}{0.8pt}
    \begin{tabularx}{0.98\linewidth}{L|CC|CCC|CCC}
        \toprule
        & \multicolumn{2}{c|}{Negative-label Test} & \multicolumn{3}{c|}{Distractor Test} & \multicolumn{3}{c}{Classic} \\
        \cmidrule(lr){2-3} \cmidrule(lr){4-6} \cmidrule(lr){7-9}
        \centering Method & {\footnotesize NMN} $\downarrow$ & {\footnotesize PCCN $\uparrow$} &
        {\footnotesize CntP $\uparrow$} & {\footnotesize CntR $\uparrow$} &
        {\footnotesize CntF1 $\uparrow$} & {\footnotesize GAME(1) $\downarrow$} &
        {\footnotesize MAE $\downarrow$} & {\footnotesize RMSE $\downarrow$} \\
        \midrule
        \midrule
        ZSC~\cite{DBLP:conf/cvpr/XuL0RS23} {\footnotesize (CVPR '23)} & 0.98 & 51.94 & 0.49 & 0.75 & 0.54 & 60.73 & 25.27 & 91.33 \\
        CounTX~\cite{AminiNaieni23} {\footnotesize (BMVC '23)} & 0.88 & 69.21 & 0.63 & 0.67 & 0.58 & 49.21 & 17.15 & 65.93 \\
        CLIP-Count~\cite{DBLP:conf/mm/JiangLC23} {\footnotesize (ACM MM '23)} & 1.24 & 47.90 & 0.48 & 0.68 & 0.52 & 61.80 & 18.82 & 66.31 \\
        VLCounter~\cite{DBLP:conf/aaai/KangMKH24} {\footnotesize (AAAI '24)} & 1.08 & 62.60 & 0.51 & 0.74 & 0.55 & 57.13 & 18.09 & \underline{65.16}\bronze\mbox{} \\
        TFPOC~\cite{10483595} {\footnotesize (WACV '24)} & 0.67 & 61.35 & 0.71 & 0.74 & 0.65 & 45.29 & 32.67 & 109.87 \\
        DAVE~\cite{Pelhan_2024_CVPR} {\footnotesize (CVPR '24)} & \textbf{0.13}\gold & \textit{95.26}\silver & 0.73 & 0.67 & \underline{0.66}\bronze & 38.56 & \underline{17.06}\bronze & \textit{56.60}\silver\mbox{} \\
        PseCo~\cite{DBLP:conf/cvpr/HuangD0ZS24} {\footnotesize (CVPR '24)} & 0.92 & 53.42 & 0.50 & 0.75 & 0.55 & 60.01 & 26.92 & 108.06 \\
        %GroundingREC~\cite{10656642} {\footnotesize (CVPR '24)}$^*$ & 0.33 & 94.17 & 0.88 & 0.81 & 0.81 & 28.52 & 21.07 & 87.57 \\ % results trained on their dataset
        GroundingREC~\cite{10656642} {\footnotesize (CVPR '24)} & \underline{0.37}\bronze & \underline{93.86}\bronze & \textbf{0.90}\gold & \underline{0.79}\bronze & \textbf{0.83}\gold & \textbf{27.33}\gold & 21.40 & 88.40 \\ % results model trained on FSC
        UPC~\cite{DBLP:conf/aaai/0018C24} {\footnotesize (AAAI '24)} & 1.03 & 62.99 & \textit{0.81}\silver & \textit{0.81}\silver & \textit{0.79}\silver & \textit{28.62}\silver & \textit{15.26}\silver & \textbf{55.71}\gold\mbox{} \\
        CountGD~\cite{DBLP:journals/corr/abs-2407-04619} {\footnotesize (NeurIPS '24)} & \textit{0.14}\silver & \textbf{95.41}\gold & \underline{0.77}\bronze & \textbf{0.84}\gold & \textit{0.79}\silver & \underline{32.49}\bronze & \textbf{12.67}\gold & 67.68 \\
        \bottomrule 
    \end{tabularx}

    %* usando il loro dataset grande; $\$$ usando il loro fsc
\end{table*}

%\subsubsection{Datasets}

%\subsubsection{Probed Methods}
We evaluate ten state‑of‑the‑art open-world text-guided CAC methods: ZSC~\cite{DBLP:conf/cvpr/XuL0RS23}, CounTX~\cite{AminiNaieni23}, CLIP‑Count~\cite{DBLP:conf/mm/JiangLC23}, VLCounter~\cite{DBLP:conf/aaai/KangMKH24}, TFPOC~\cite{10483595}, DAVE~\cite{Pelhan_2024_CVPR}, PseCo~\cite{DBLP:conf/cvpr/HuangD0ZS24}, GroundingREC~\cite{10656642}, UPC~\cite{DBLP:conf/aaai/0018C24}, and CountGD~\cite{DBLP:journals/corr/abs-2407-04619}. 
Among these approaches, ZSC, CounTX, CLIP‑Count, and VLCounter adopt architectures that directly regress density maps by fine‑tuning CLIP and conditioning the density‑map prediction on the CLIP embedding of the queried object class. In contrast, TFPOC, DAVE, PseCo, GroundingREC, UPC, and CountGD follow a more detection‑oriented two‑stage paradigm, where objects are first detected in the image and then filtered based on the input textual prompt. For more details, we refer to Sec.~\ref{sec:sec:related-work-sota} and to the original papers associated with the methods discussed above.
%\subsubsection{Implementation Details}

We relied on the original implementations and pre‑trained models released by the respective authors, preserving their image pre‑processing pipelines, hyperparameter configurations, and prompting strategies. Only minor adjustments were required to ensure that the DAVE inference procedure operated correctly within our benchmark. Details on these modifications are provided in~\ref{sec:appendix-implementation}. Another minor change concerns GroundingREC: we retrained the model by filtering the original REC‑8K dataset to include only images from FSC‑147, ensuring a fair comparison across all models. 

Finally, we report results not only for the negative-label and distractor tests, but also with respect to standard counting metrics, namely MAE, RMSE, and GAME(1). For the latter, we set $L=1$ to align with the patch partitioning used in the distractor test, while additional results for $L=2$ and $L=3$ are reported in~\ref{sec:additional-results}.

\begin{table*}[!t]
\centering
    \caption{\textbf{Results on our \datasetacron\ dataset.} We report the performance of the ten SOTA methods analyzed in this work using our \benchmarkacron{} test suite (negative-label and distractor tests), along with standard counting-specific evaluation metrics. Best, second-best, and third-best results are indicated with \gold, \silver, and \bronze, and formatted using \textbf{bold}, \textit{italic}, and \underline{underline}, respectively.}
    \vspace{3pt}
    \label{tab:results-MUCCA}%
    \newcolumntype{L}{>{\arraybackslash}m{.27\linewidth}}%
    \newcolumntype{C}{>{\centering\arraybackslash}X}
    %\tiny%
    \setlength{\tabcolsep}{0.8pt}
    \begin{tabularx}{0.98\linewidth}{l|CC|CCC|CCC}
    \toprule
    & \multicolumn{2}{c|}{Negative-label Test} & \multicolumn{3}{c|}{Distractor Test} & \multicolumn{3}{c}{Classic} \\
    \cmidrule(lr){2-3} \cmidrule(lr){4-6} \cmidrule(lr){7-9}
    \centering Method & {\footnotesize NMN} $\downarrow$ & {\footnotesize PCCN $\uparrow$} &
    {\footnotesize CntP $\uparrow$} & {\footnotesize CntR $\uparrow$} &
    {\footnotesize CntF1 $\uparrow$} & {\footnotesize GAME(1) $\downarrow$} &
    {\footnotesize MAE $\downarrow$} & {\footnotesize RMSE $\downarrow$} \\
    \midrule
    \midrule
    ZSC~\cite{DBLP:conf/cvpr/XuL0RS23} {\footnotesize (CVPR '23)} & 0.95 & \textit{54.50}\silver & 0.49 & 0.76 & 0.48 & 34.43 & 30.30 & 32.58 \\
    CounTX~\cite{AminiNaieni23} {\footnotesize (BMVC '23)} & 1.25 & 46.50 & 0.34 & 0.87 & 0.44 & 37.22 & 32.41 & 36.02 \\
    CLIP-Count~\cite{DBLP:conf/mm/JiangLC23} {\footnotesize (ACM MM '23)} & 0.95 & 44.00 & 0.47 & \textbf{0.89}\gold & \underline{0.54}\bronze & 26.13 & 22.97 & 25.44 \\
    VLCounter~\cite{DBLP:conf/aaai/KangMKH24} {\footnotesize (AAAI '24)} & 0.66 & 49.50 & 0.47 & 0.80 & 0.53 & 26.29 & 22.69 & \underline{25.17}\bronze\mbox{} \\
    TFPOC~\cite{10483595} {\footnotesize (WACV '24)} & 0.44 & 45.00 & \underline{0.52}\bronze & 0.72 & 0.51 & 25.33 & 22.42 & 26.20 \\
    % DAVE & 0.22 & 48.50 & 0.33 & -0.18 & 0.37 & 26.39 & 30.86 & 21.93 & 25.14 \\
    DAVE~\cite{Pelhan_2024_CVPR} {\footnotesize (CVPR '24)} & \underline{0.21}\bronze & 49.00 & \underline{0.52}\bronze & 0.65 & 0.46 & \underline{24.31}\bronze & \underline{22.05}\bronze & 25.43 \\ % ReLU on density map
    %\rowcolor{yellow}
    %PseCo & 1.05 & 44.50 & 0.15 & 0.26 & 0.19 & 56.55 & 29.18 & 32.13 \\
    PseCo~\cite{DBLP:conf/cvpr/HuangD0ZS24} {\footnotesize (CVPR '24)} & 1.07 & 44.00 & 0.38 & 0.75 & 0.44 & 36.52 & 30.19 & 32.98 \\
    GroundingREC~\cite{10656642} {\footnotesize (CVPR '24)} & \textit{0.18}\silver & \textbf{91.00}\gold & \textbf{0.71}\gold & \underline{0.79}\bronze & \textbf{0.72}\gold & \textbf{11.57}\gold & \textbf{9.58}\gold & \textbf{12.36}\gold\mbox{} \\
    UPC~\cite{DBLP:conf/aaai/0018C24} {\footnotesize (AAAI '24)} & 1.07 & 50.50 & \textit{0.53}\silver & \underline{0.79}\bronze & 0.53 & 28.32 & 25.14 & 28.24 \\
    CountGD~\cite{DBLP:journals/corr/abs-2407-04619} {\footnotesize (NeurIPS '24)} & \textbf{0.09}\gold & \underline{53.00}\bronze & 0.50 & \textit{0.83}\silver & \textit{0.58}\silver & \textit{21.43}\silver & \textit{18.39}\silver & \textit{20.89}\silver\mbox{} \\
    \bottomrule
    \end{tabularx}
\end{table*}

\begin{figure*}[!t]
    \centering
    \includegraphics[width=0.80\linewidth]{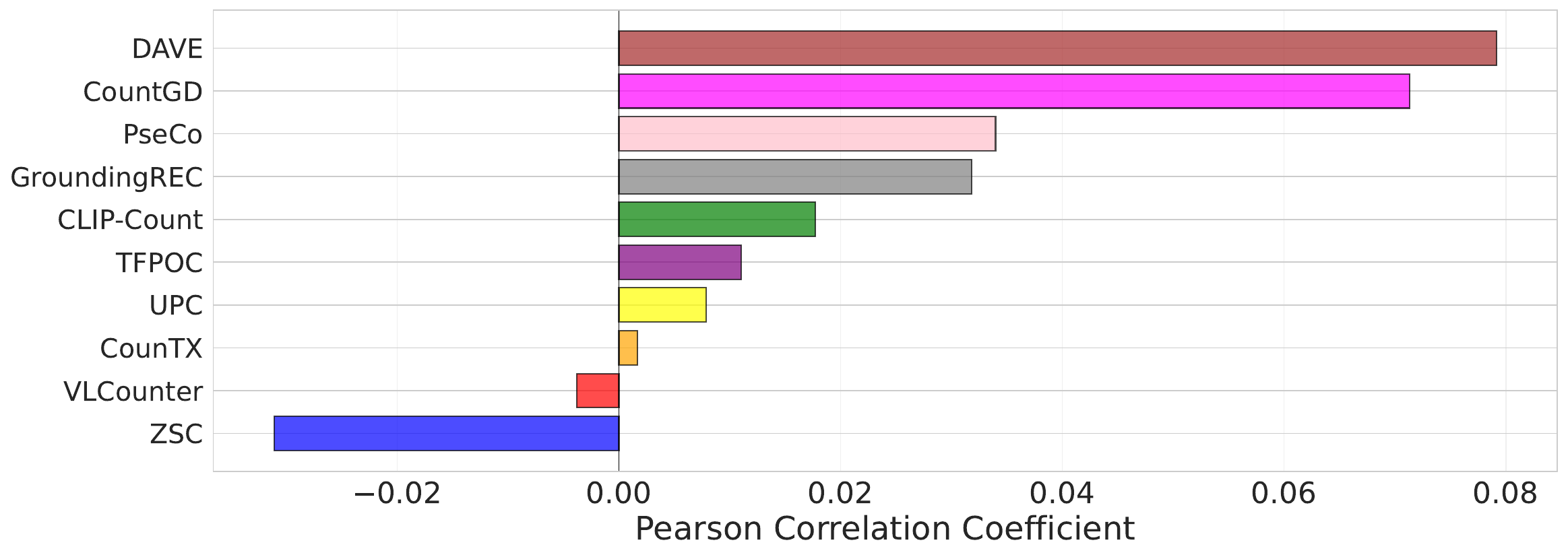}
    \includegraphics[width=0.80\linewidth]{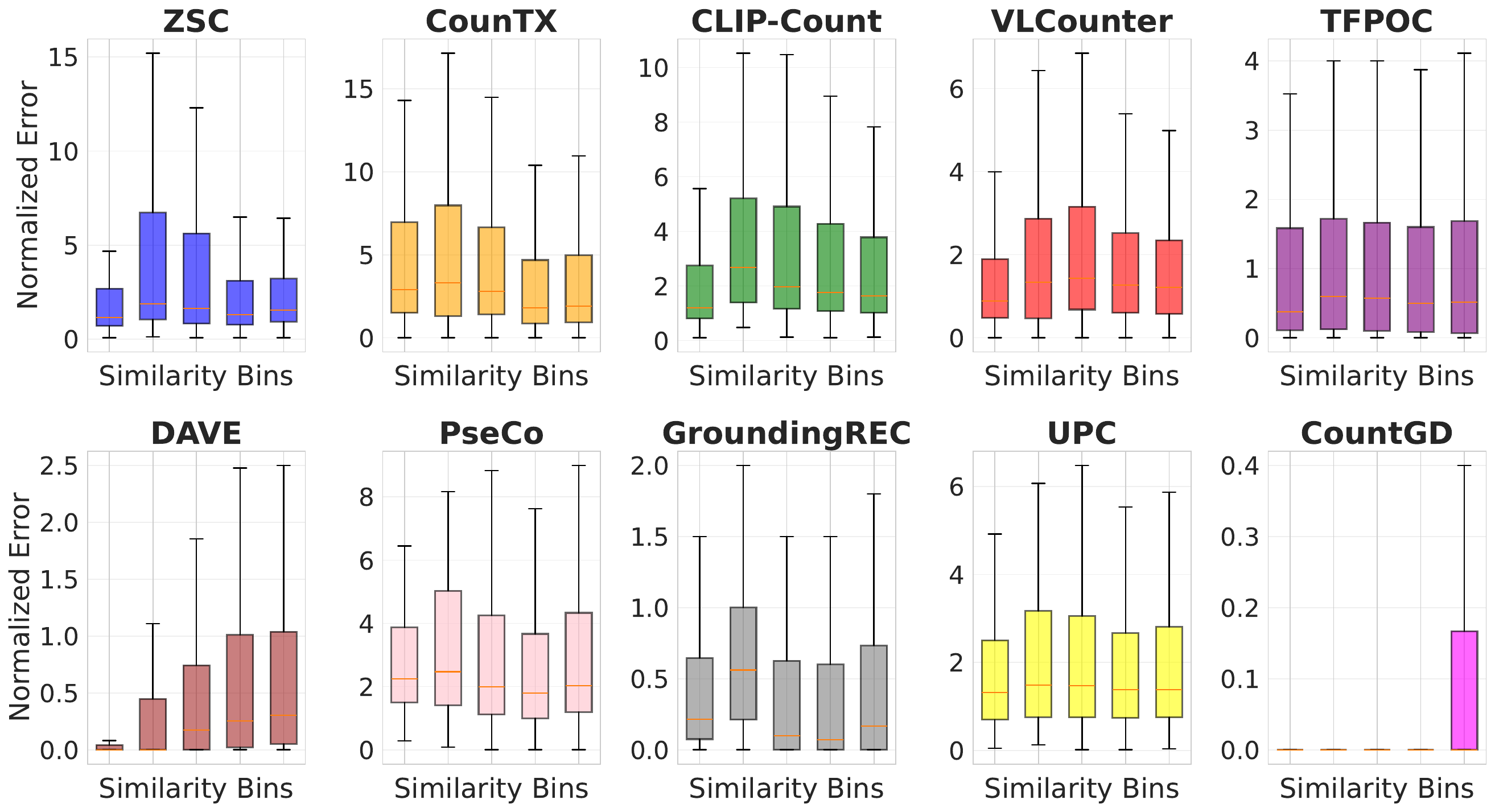}
    \caption{\textbf{Effect of textual prompt semantic similarity in the negative-label test.}
    (a) Pearson correlation between semantic similarity (CLIP cosine similarity) and normalized counting error for each evaluated model.
    (b) Quartile distributions of normalized counting errors grouped into five equal-width semantic similarity bins.
    For each negative category in an image, the counting error is given by the model prediction (ground-truth count equal to zero) and normalized by the ground-truth count of the positive category with maximum semantic similarity.
    Box plots report first and third quartiles, median, and mean across samples within each bin.}
    \label{fig:impact-semantic-similarity-bins}
\end{figure*}

\subsection{Results}

\subsubsection{Quantitative Results on FSC-147}
Table~\ref{tab:results-FSC147-test-set} and Tab.~\ref{tab:results-FSC147-val-set} report the results for the test and validation splits of the FSC-147 dataset~\cite{DBLP:conf/cvpr/RanjanSNH21}, respectively. As shown, although the methods achieve strong performance on standard counting error measures (MAE and RMSE), their behavior on the \benchmarkacron{} metrics is considerably more heterogeneous. Notably, the negative-label test on the test split reveals that one-stage methods such as ZSC~\cite{DBLP:conf/cvpr/XuL0RS23}, CounTX~\cite{AminiNaieni23}, CLIP-Count~\cite{DBLP:conf/mm/JiangLC23}, and VLCounter~\cite{DBLP:conf/aaai/KangMKH24} tend to produce an average negative count comparable to the ground-truth count of the queried class ($\text{NMN} \approx 1$), with ZSC, CLIP-Count, and VLCounter even yielding $\text{NMN} > 1$ (i.e., yelding a count for the negative classes higher than the ground truth count for the positive ones). This trend is further reflected in the PCCN metric: for instance, CLIP-Count produces an estimate closer to the ground truth only 38\% of the time on the test set.
Some detection-based methods also exhibit the same limitation, such as PseCo~\cite{DBLP:conf/cvpr/HuangD0ZS24} and UPC~\cite{DBLP:conf/aaai/0018C24}. In contrast, the other detection-based approaches---DAVE~\cite{Pelhan_2024_CVPR}, CountGD~\cite{DBLP:journals/corr/abs-2407-04619}, and GroundingREC~\cite{10656642}---achieve the best, second-best, and third-best performance on the negative-label test, respectively. A similar trend is observed on the validation split, with only minor variations in PCCN between DAVE and CountGD.

Slightly different trends can be observed on the distractor test conducted with the mosaic implementation. Specifically, although the top three models remain DAVE, GroundingREC, and CountGD in most cases, UPC also achieves competitive performance, occasionally reaching the second-best score on both the test and validation splits. It is worth noting that performance gaps between the best- and worst-performing methods become more evident when considering the CntP metric: values range from 0.50 for ZSC and VLCounter to 0.89 for GroundingREC (test split), whereas differences in CntR are less pronounced. This suggests that, while most methods can count the correct class reasonably well, their robustness to distractor images within mosaics varies significantly.

Interestingly, more recent methods do not necessarily achieve better performance on the \benchmarkacron{} metrics, indicating that this aspect of evaluation remains largely overlooked. For example, UPC attains near–state-of-the-art results on classic counting metrics but exhibits catastrophic performance on the negative-label test. Conversely, DAVE appears to offer the most balanced performance when comparing traditional counting metrics with those introduced in \benchmarkacron{}.

\subsubsection{Quantitative Results on \datasetacron}
Table~\ref{tab:results-MUCCA} reports the results on our \datasetacron{} dataset, which introduces novel and unique challenges for all the evaluated methods. GroundingREC~\cite{10656642} emerges as the overall best-performing approach, achieving the top results across almost all metrics and exhibiting an excellent balance between strong performance on both the classic counting metrics and those introduced by \benchmarkacron{}.
DAVE~\cite{Pelhan_2024_CVPR} and CountGD~\cite{DBLP:journals/corr/abs-2407-04619}, which were among the best-performing methods in the previous experiments, also show competitive results, although they generally fall behind GroundingREC on \benchmarkacron{}. Moreover, their standard counting error can occasionally double in this evaluation setting.
The behavior of ZSC~\cite{DBLP:conf/cvpr/XuL0RS23}, CounTX~\cite{AminiNaieni23}, CLIP-Count~\cite{DBLP:conf/mm/JiangLC23}, PseCo~\cite{DBLP:conf/cvpr/HuangD0ZS24}, and UPC~\cite{DBLP:conf/aaai/0018C24} is consistent with what was observed earlier: all these methods still produce $\text{NMN} \approx 1$ and sometimens even $NMN > 1$, indicating a persistent inability to handle negative-label cases effectively.

\subsubsection{Impact of Textual Prompt Semantic Similarity}
In order to investigate whether counting errors are influenced by the semantic similarity between target and non-target categories, we analyze model behavior in the \textit{negative-label} test as a function of textual prompt similarity. Specifically, we compute the semantic similarity between two object categories as the cosine similarity between their CLIP~\cite{DBLP:conf/icml/RadfordKHRGASAM21} text embeddings---we adopt the CLIP ViT-B/32 textual encoder and prepend each category label with the prompt template ``\textit{a photo of}''.
Then, for each image and each tested negative category, we associate a similarity score defined as the maximum cosine similarity between the negative category and the set of positive categories present in that image. This choice reflects the semantic proximity of the negative category to the visual content of the image, focusing on the most challenging case for each negative prompt.
%In order to assess if a model tends to wrongly count negative categories that are more semantically similar to the positive ones present in each image, we computed the semantic similarity between two categories as the CLIP~\cite{DBLP:conf/icml/RadfordKHRGASAM21} cosine similarity between the text embeddings of their corresponding labels. We used CLIP ViT B/32 textual encoder, and used as a prefix for each label the string "a photo of ". 

\begin{figure*}[!t]
    \centering
    \includegraphics[width=0.95\linewidth]{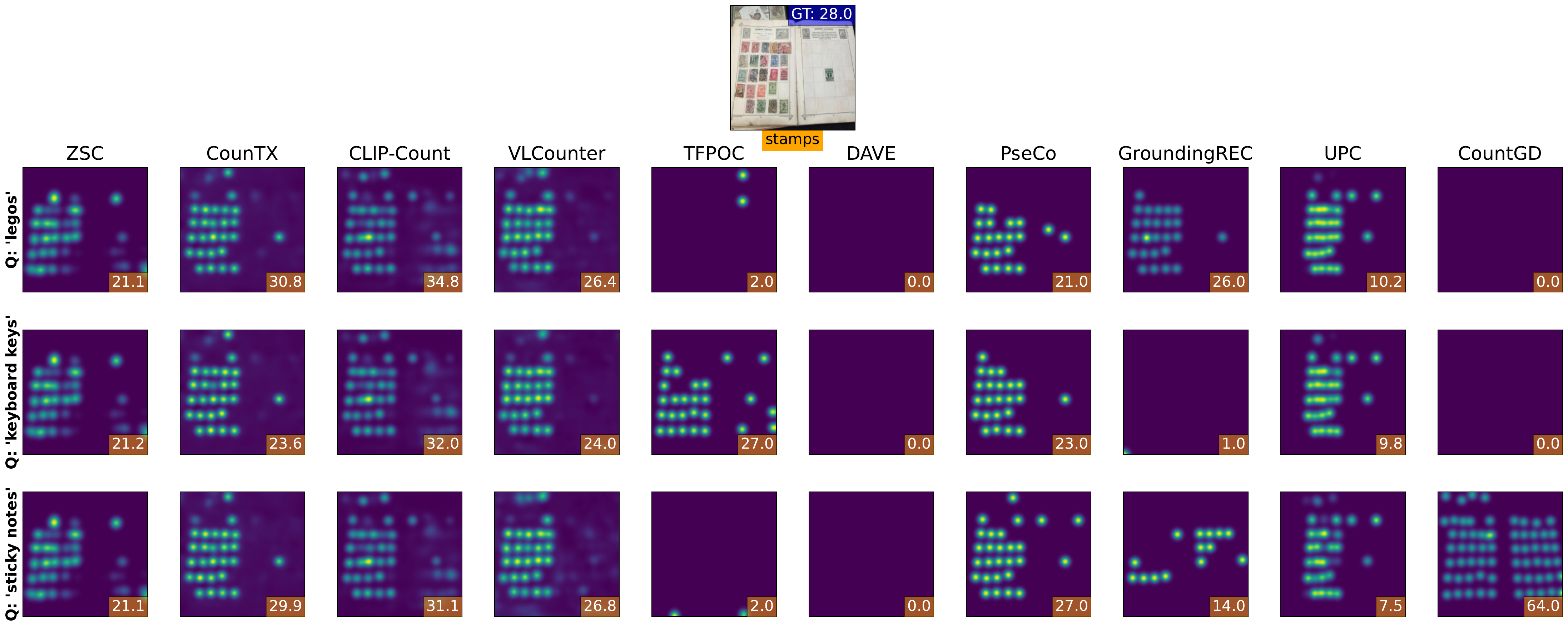}
    \par\vspace{0.8cm}
    \includegraphics[width=0.95\linewidth]{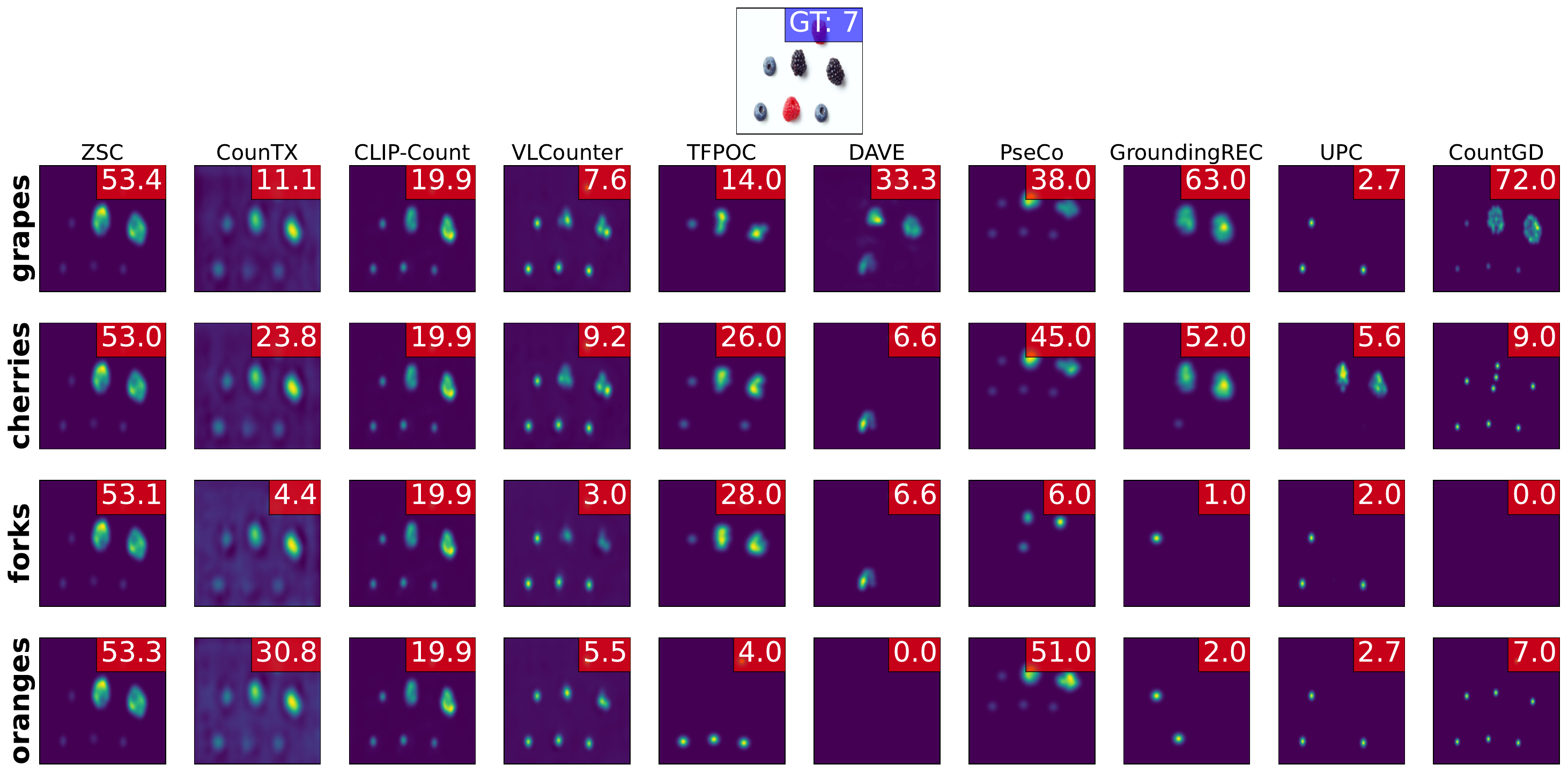}
    %\caption{\textbf{Qualitative Results on the Negative test.} TODO.}
    \caption{\textbf{Qualitative results for the negative-label test.} We present sample images from both the FSC-147 and \datasetacron{} datasets, alongside the density maps generated by the evaluated text-guided CAC models when queried with absent object categories (negative prompts). %While robust methods like DAVE and CountGD correctly suppress activations and predict near-zero counts, most other approaches erroneously highlight salient but irrelevant objects, revealing a significant lack of semantic grounding. The intrinsically multi-class nature of \datasetacron{} (bottom rows) further exacerbates this issue, occasionally causing even the best-performing models to output spurious counts.
    }
    \label{fig:qualitative-results-negative-test}
\end{figure*}

\begin{figure*}[!t]
    \centering
    \includegraphics[width=0.95\linewidth]{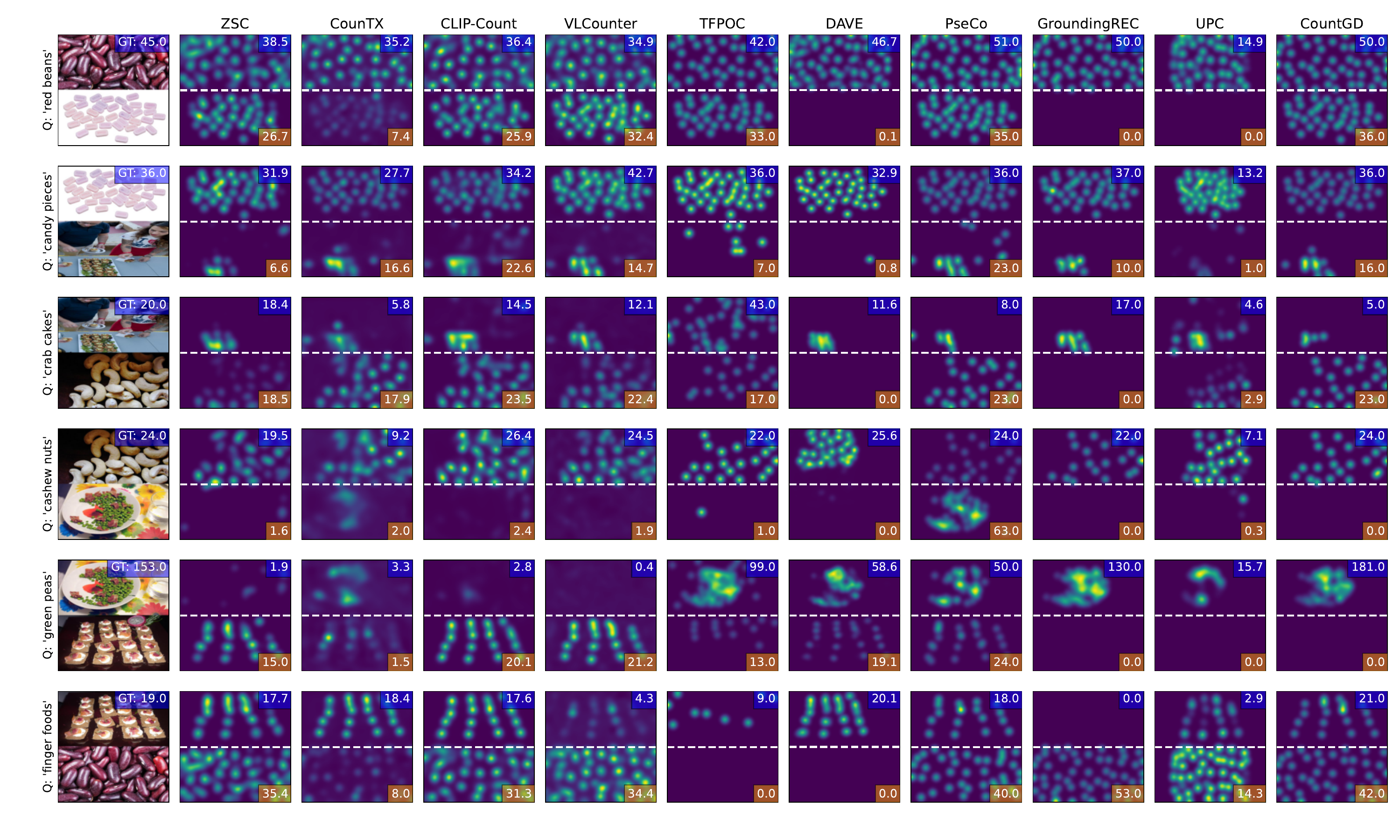}
    \par\vspace{0.3cm}
    \includegraphics[width=0.95\linewidth]{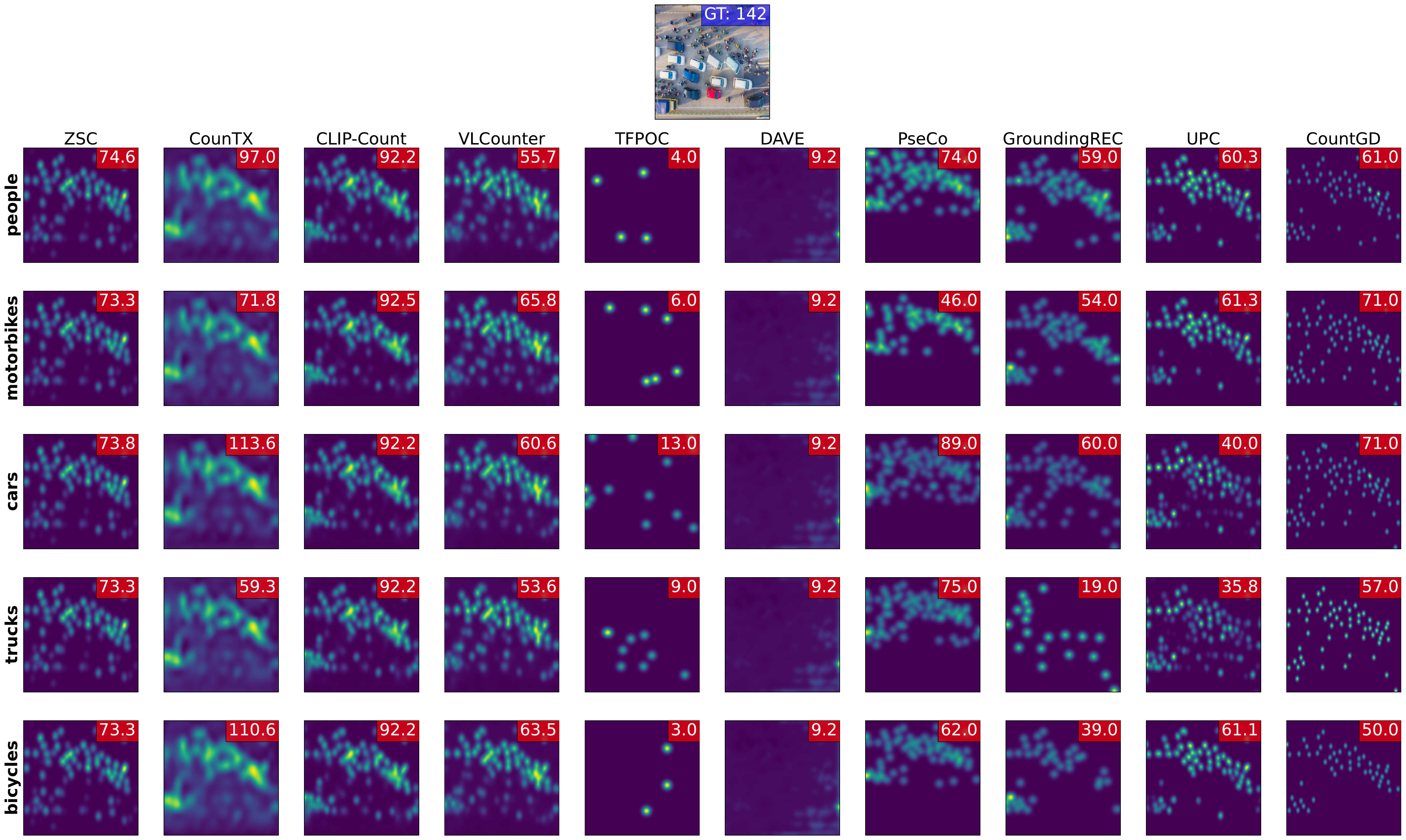}
    %\caption{\textbf{Qualitative Results on the Distractor test.} TODO.}
    \caption{\textbf{Qualitative results for the distractor test.} We showcase the behavior of various models when tasked with counting a target class in the presence of confusing, co-occurring object categories. The examples include synthetic mosaics constructed from single-class FSC-147 images (top rows) and real-world multi-class scenes from the \datasetacron{} dataset (bottom rows). %As evidenced by the predicted density maps, many state-of-the-art models struggle to selectively attend to the target category, frequently accumulating spurious counts on visually salient distractors. This highlights the challenge current architectures face in properly disentangling multiple object classes based solely on textual prompts.
    }
    \label{fig:qualitative-results-distractor-test}
\end{figure*}

Figure~\ref{fig:impact-semantic-similarity-bins}(a) reports, for each evaluated model, the Pearson correlation between the counting error and the semantic similarity score computed over the images of the \datasetacron{} dataset. This correlation provides a compact summary of whether counting errors tend to increase when negative categories are more semantically similar to the positive ones present in the image. 
Instead, Fig.~\ref{fig:impact-semantic-similarity-bins}(b) provides a more detailed, sample-level view of this relationship. Samples are grouped into five equal-width semantic similarity bins defined over the range of cosine similarity values, and the distribution of the normalized counting error is reported within each bin for all assessed models. Each sample corresponds to an image–negative-category pair. For each sample, the counting error is given by the prediction produced by the model for a negative category (whose ground-truth count is zero), normalized by the ground-truth count of the most semantically similar positive category present in the same image. This normalization enables comparisons across images characterized by different object densities. The inter-quartile ranges shown in Fig.~\ref{fig:impact-semantic-similarity-bins}(b) illustrate how both the magnitude and the variability of counting errors evolve across increasing levels of semantic similarity, complementing the correlation analysis in Fig.~\ref{fig:impact-semantic-similarity-bins}(a) with a finer-grained, distributional perspective.

Overall, the results indicate that semantic similarity between negative and positive categories can have a measurable impact on counting performance, although the strength of this effect varies across models. From a comparative perspective, several approaches exhibit a clearer positive correlation between semantic similarity and normalized counting error, as well as a more pronounced increase in error across similarity bins. In contrast, other models display flatter error distributions and weaker correlations, suggesting a reduced sensitivity to semantic proximity between negative and positive categories.
In particular, models such as CLIP-Count~\cite{DBLP:conf/mm/JiangLC23}, CounTX~\cite{AminiNaieni23}, and VLCounter~\cite{DBLP:conf/aaai/KangMKH24} show a more evident dependency on semantic similarity, with normalized counting errors increasing consistently as similarity grows. Conversely, approaches such as DAVE~\cite{Pelhan_2024_CVPR}, CountGD~\cite{DBLP:journals/corr/abs-2407-04619}, and UPC~\cite{DBLP:conf/aaai/0018C24} appear comparatively more stable across similarity levels, indicating a higher robustness to semantic ambiguity in negative prompts.

%Figure~\ref{fig:impact-semantic-similarity-bins} a) shows the correlation between the counting error made by each model on the images from \benchmarkacron{} dataset when probing negative classes that are more similar to the positive categories; while Fig.~\ref{fig:impact-semantic-similarity-bins} b) reports the detail of the averaged counting error obtained by each of the assessed models at different semantic similarity levels.
%The semantic similarity bins are defined as equal-width ranges of the cosine similarity scores.
%The similarity score adopted for each negative category tested on each image is the score obtained with the most similar positive class for that same image.
%Figure~\ref{fig:impact-semantic-similarity-bins} reports the inter-quartile ranges for 5 equal-width bins, and the counting error of each sample is the prediction provided by each assessed model for each negative category. That counting error is normalized by the ground truth count of the most similar class.

\subsubsection{Qualitative Results}
Figure~\ref{fig:qualitative-results-negative-test} shows qualitative examples from the negative-label test. We include images from both FSC-147~\cite{DBLP:conf/cvpr/RanjanSNH21} and \datasetacron{}, used as inputs to the ten SOTA methods under evaluation, each paired with several negative classes. For FSC-147, which contains one annotated object category per image, only DAVE~\cite{Pelhan_2024_CVPR}, CountGD~\cite{DBLP:journals/corr/abs-2407-04619}, and---to a lesser extent---GroundingREC~\cite{10656642} produce density maps that consistently predict zero (or close-to-zero) instances when queried with an absent class, in agreement with the quantitative results. 
In contrast, the remaining methods often output a number of instances comparable to, or even higher than, the ground-truth count of the true class, revealing a clear inability to handle negative prompts. 
On the other hand, \datasetacron{} represents a more challenging scenario: even the three best-performing models---DAVE, GroundingREC, and CountGD---sometimes generate density maps that highlight clear, noticeable errors.

%%%%%%%%%%%%%%%%%%%%%%%%%%%%%%%%%%%%%%%%%%%%%%%%%%%%%%%%%%%%%%%%%%%%%%%%%%%%%%%%%%%%%%%%%%%%%%
\section{Conclusions}
\label{sec:conclusions}
In this paper, we addressed a key limitation of current benchmarks for evaluating open‑world text‑guided class‑agnostic counting (CAC), namely their inability to assess whether models correctly identify \emph{which} object category is specified by a textual prompt. Our analysis shows that many state‑of‑the‑art approaches exhibit a weak alignment between textual semantics and visual representations, often resulting in hallucinated counts driven by dominant object categories in the scene rather than by the actual query.

We argue that these failures arise from the use of class‑specific evaluation metrics focused solely on counting accuracy, as well as from the limitations of existing CAC datasets, which predominantly contain single‑category images. To address this gap, we introduced \benchmarkacron{} (\benchmarkname), a new test suite based on the \textit{negative‑label} and \textit{distractor} tests, explicitly designed to probe robustness to misleading prompts and multi‑category scenes. In addition, we presented \datasetacron{} (\datasetname), a multi‑category dataset of real‑world images created to support these evaluations.

To thoroughly examine the effectiveness of our contributions, we conducted an extensive experimental study covering ten state-of-the-art open‑world text‑guided CAC methods. Our analysis shows that although several approaches achieve strong results under standard class-specific evaluation protocols, they often struggle to correctly associate textual descriptions with their corresponding visual categories. Furthermore, the coexistence of multiple object categories within the same image turns out to be a substantial source of error for many of the evaluated techniques. We also conducted a complementary analysis to understand the origins of these failure cases. Besides qualitative examples, this analysis includes a quantitative examination of how the semantic similarity between the textual prompt and the visual content relates to model performance. In some instances, this relationship appears measurable, though its strength varies across methods. We expect that both our benchmark and dataset will provide a useful reference for future research, helping to expose the limitations of current approaches and highlighting the need for improved training pipelines or possibly revised architectural choices.

%%%%%%%%%%%%%%%%%%%%%%%%%%%%%%%%%%%%%%%%%%%%%%%%%%%%%%%%%%%%%%%%%%%%%%%%%%%%%%%%%%%%%%%%%%%%%%
\printcredits

\section*{Declaration of generative AI and AI-assisted technologies in the manuscript preparation process}
The authors acknowledge the use of AI-assisted writing tools, including ChatGPT, as well as Grammarly, to improve the clarity and readability of the manuscript. These tools were used in accordance with academic integrity principles and only for language editing purposes. All scientific content and contributions are the original work of the authors.

\section*{Declaration of competing interest}
The authors declare that they have no known competing financial interests or personal relationships that could have appeared to influence the work reported in this paper.

\section*{Acknowledgements}
This work was partially supported by 
Spoke 8, Tuscany Health Ecosystem (THE) Project (CUP B83C22003930001), funded by the National Recovery and Resilience Plan (NRRP), within the NextGeneration Europe (NGEU) Program; Horizon Europe Research \& Innovation Programme under Grant agreement N. 101092612 (Social and hUman ceNtered XR - SUN project); PNRR - M4C2 - Investimento 1.3, Partenariato Esteso PE00000013 - "FAIR - Future Artificial Intelligence Research" - Spoke 1 "Human-centered AI", funded by European Union - NextGenerationEU; ITSERR - ITalian Strengthening of the Esfri Ri Resilience (CUP B53C22001770006), also funded by the European Union via NextGenerationEU, 
the FoReLab and CrossLab projects (Departments of Excellence) funded by the Italian Ministry of Education and Research (MUR), and the NVIDIA Academic Grants Program 2026.

\section*{Data availability}
The datasets are publicly available. The \datasetacron{} dataset can be freely downloaded from Zenodo at \href{https://zenodo.org/records/19231375}{zenodo.org/\-records/\-19231375}.

\appendix
\section{Generalizing Counting Precision and Recall}
\label{sec:conting-prec-rec}
%- Formalizzazione che spiega come si torna dal generale al particolare (mosaico)
%- Risultati empirici che mostrano che le differenze (o correlazioni) tra le metriche
In this appendix, we formally establish the relationship between the generalized definition of counting precision and recall concerning the \textit{distractor} test introduced in this paper and the original mosaic-based formulation proposed in~\cite{DBLP:conf/wacv/CiampiMP0AF25}. In particular, we show that the latter can be recovered as a special case of the proposed patch-based formulation under mild and intuitive assumptions. This derivation clarifies the continuity between the two approaches and confirms that the proposed metrics constitute a proper generalization. 
%rather than a fundamentally different evaluation protocol.
%In this section, we present and fill the gap between the formulations of counting precision and recall reported in this paper and those reported in~\cite{DBLP:conf/wacv/CiampiMP0AF25}. Specifically, we show how the old mosaic-based formulation can be derived from the novel one by performing straightforward reduction steps.

\begin{table*}[!t]
\caption{\textbf{Comparison between the proposed patch-based counting precision/recall and the original mosaic-based formulation}. Results are reported on the FSC-147 test split. Values outside parentheses correspond to the proposed metrics, while values in parentheses denote the original mosaic-based definitions introduced in~\cite{DBLP:conf/wacv/CiampiMP0AF25}.}
\label{tab:new-vs-old-metrics}
\vspace{3pt}
\newcolumntype{C}{>{\centering\arraybackslash}X}
\setlength{\tabcolsep}{0.8pt}
\centering
\begin{tabularx}{0.68\linewidth}{lCCC}
\toprule
Model & Precision & Recall & F1-Score \\
\midrule
\midrule
ZSC~\cite{DBLP:conf/cvpr/XuL0RS23} {\footnotesize (CVPR '23)} 
& 0.50 {\scriptsize (0.51)} 
& 0.82 {\scriptsize (0.83)} 
& 0.57 {\scriptsize (0.58)} \\

CounTX~\cite{AminiNaieni23} {\footnotesize (BMVC '23)} 
& 0.66 {\scriptsize (0.69)} 
& 0.72 {\scriptsize (0.71)} 
& 0.63 {\scriptsize (0.63)} \\

CLIP-Count~\cite{DBLP:conf/mm/JiangLC23} {\footnotesize (ACM MM '23)} 
& 0.49 {\scriptsize (0.49)} 
& 0.75 {\scriptsize (0.76)} 
& 0.55 {\scriptsize (0.55)} \\

VLCounter~\cite{DBLP:conf/aaai/KangMKH24} {\footnotesize (AAAI '24)} 
& 0.50 {\scriptsize (0.51)} 
& 0.78 {\scriptsize (0.78)} 
& 0.57 {\scriptsize (0.57)} \\

TFPOC~\cite{10483595} {\footnotesize (WACV '24)} 
& 0.68 {\scriptsize (0.69)} 
& 0.83 {\scriptsize (0.85)} 
& 0.69 {\scriptsize (0.69)} \\

DAVE~\cite{Pelhan_2024_CVPR} {\footnotesize (CVPR '24)} 
& 0.78 {\scriptsize (0.84)} 
& 0.73 {\scriptsize (0.80)} 
& 0.72 {\scriptsize (0.79)} \\

PseCo~\cite{DBLP:conf/cvpr/HuangD0ZS24} {\footnotesize (CVPR '24)} 
& 0.53 {\scriptsize (0.53)} 
& 0.85 {\scriptsize (0.87)} 
& 0.61 {\scriptsize (0.61)} \\

GroundingREC~\cite{10656642} {\footnotesize (CVPR '24)} 
& 0.89 {\scriptsize (0.89)} 
& 0.80 {\scriptsize (0.81)} 
& 0.82 {\scriptsize (0.83)} \\

UPC~\cite{DBLP:conf/aaai/0018C24} {\footnotesize (AAAI '24)} 
& 0.80 {\scriptsize (0.80)} 
& 0.82 {\scriptsize (0.83)} 
& 0.78 {\scriptsize (0.79)} \\

CountGD~\cite{DBLP:journals/corr/abs-2407-04619} {\footnotesize (NeurIPS '24)} 
& 0.74 {\scriptsize (0.74)} 
& 0.86 {\scriptsize (0.88)} 
& 0.78 {\scriptsize (0.79)} \\

\bottomrule
\end{tabularx}
\end{table*}

\paragraph{Mosaic setting as a special case}
In the mosaic-based evaluation protocol in~\cite{DBLP:conf/wacv/CiampiMP0AF25}, each test image is synthetically constructed by vertically concatenating two images: a \emph{positive} image $I^{\text{pos}}$ containing instances of the target class, and a \emph{negative} image $I^{\text{neg}}$ containing instances of a different class. As a result, the image can be naturally decomposed into two disjoint spatial regions (or patches), which we denote as $g_1$ (upper patch) and $g_2$ (lower patch). Let $\tilde{c}_1$ and $\tilde{c}_2$ denote the ground-truth counts of the target class in the two patches. By construction, $\tilde{c}_1 > 0$ and $\tilde{c}_2 = 0$, i.e., all true instances of the target class are confined to the upper region, while the lower region contains only distractor objects.
%In particular, when specializing the new formulation to the mosaic one, we are performing the following assumptions: (i) we are dividing the image into only two patches, an upper patch (corresponding to $I^\text{pos}$ of the mosaic), and a lower patch (corresponding to $I^\text{neg}$ of the mosaic); (ii) $I^\text{neg}$ only contributes to false positives, given that we know a-priori -- from the mosaic construction process -- that there are no correct instances in that part of the image.

Let $c_1$ and $c_2$ be the corresponding predicted counts produced by the model in the two patches. Following the general definitions introduced in the main paper, the patch-wise contributions of true positives (TP), false positives (FP), and false negatives (FN) are defined as follows.

\paragraph{Upper patch ($g_1$)}
\begin{equation}
\begin{cases}
\text{TP}_1 = \min(c_1, \tilde{c}_1), \\
\text{FP}_1 = \max(0, c_1 - \tilde{c}_1), \\
\text{FN}_1 = \max(0, \tilde{c}_1 - c_1).
\end{cases}
\end{equation}

\paragraph{Lower patch ($g_2$)}
%Since no ground-truth instances of the target class are present in the negative region ($\tilde{c}_2 = 0$), the corresponding contributions simplify to:
\begin{equation}
\begin{cases}
\text{TP}_2 = \min(c_2, \tilde{c}_2) = 0, \\
\text{FP}_2 = \max(0, c_2 - \tilde{c}_2) = c_2, \\
\text{FN}_2 = \max(0, \tilde{c}_2 - c_2) = 0.
\end{cases}
\end{equation}

%In particular, in the case of the mosaic, $L=2$. Supposing that $l=1$ and $l=2$ correspond to the upper and lower patches, respectively, the contributions of TPs, FPs, and FNs are as follows.

%For the upper patch:
%\begin{equation}
%\begin{cases}
%\text{TP}_1 = \min(c_1, \tilde{c}_1) \\
%\text{FP}_1 = \max(0, c_1 - \tilde{c}_1) \\
%\text{FN}_1 = \max(0, \tilde{c}_{i,l} - c_{i,l})
%\end{cases}
%\end{equation}

%For the lower patch, for which $\tilde{c}_2 = 0$:
%\begin{equation}
%\begin{cases}
%  \text{TP}_2 = \min(c_2, \tilde{c}_2) = 0\\
%  \text{FP}_2 = \max(0, c_2 - \tilde{c}_2) = c_2 \\
%  \text{FN}_2 = \max(0, \tilde{c}_2 - c_2) = 0
%\end{cases}
%\end{equation}

\paragraph{Image-level precision and recall}
Aggregating the contributions from the two patches, the image-level counting precision and recall are obtained as:
\begin{equation}
\begin{aligned}
\text{CntP} &= \frac{\text{TP}_1 + \text{TP}_2}{\text{TP}_1 + \text{TP}_2 + \text{FP}_1 + \text{FP}_2}
           = \frac{\min(c_1, \tilde{c}_1)}{c_1 + c_2}, \\
\text{CntR} &= \frac{\text{TP}_1 + \text{TP}_2}{\text{TP}_1 + \text{TP}_2 + \text{FN}_1 + \text{FN}_2}
           = \frac{\min(c_1, \tilde{c}_1)}{\tilde{c}_1}.
\end{aligned}
\end{equation}

The resulting expressions exactly coincide with the definitions of counting precision and recall introduced in~\cite{DBLP:conf/wacv/CiampiMP0AF25} for the mosaic-based evaluation protocol. As in the main paper, dataset-level metrics are obtained by averaging the image-level scores across all test images.

\paragraph{Empirical consistency analysis}
To further assess the practical impact of the proposed generalization, Tab.~\ref{tab:new-vs-old-metrics} reports a quantitative comparison between the original mosaic-based metrics and their generalized counterparts on the FSC-147 test split. The results show that the two formulations yield highly consistent performance across a wide range of state-of-the-art counting models. Minor numerical differences arise from the different aggregation strategy but do not affect the relative ranking of the evaluated methods. This empirical evidence supports the validity of the proposed formulation while highlighting its increased flexibility in handling arbitrary spatial configurations beyond the mosaic setting.

%Note that the final obtained formulas are the ones presented in \cite{DBLP:conf/wacv/CiampiMP0AF25} for the counting precision and recall on the mosaic test. 
%To better evaluate the differences empirically, we report in Table \ref{tab:new-vs-old-metrics} the new versus old precision-recall metrics for the various probed methods.

\section{Additional Results}
\label{sec:additional-results}
In this section, we extend the main experimental evaluation by analyzing model performance on the \textit{distractor} test under stricter spatial constraints. Indeed, in the main paper, the proposed patch-based counting metrics (CntP, CntR, CntF1) were computed using a grid level of $L=1$, which partitions each image into four coarse regions. Here, we report results obtained at finer spatial granularities, namely $L=2$ (16 patches) and $L=3$ (64 patches).

As expected, increasing the grid level $L$ imposes more stringent spatial alignment requirements between the predicted density maps and the ground-truth annotations. Consequently, we observe a consistent degradation in both standard counting metrics and metrics introduced with our \textit{distractor} test across all evaluated methods on the FSC-147 test set (Tab.~\ref{tab:fsc-test-level2-and-3}), the FSC-147 validation set (Tab.~\ref{tab:fsc-val-level2-and-3}), and \benchmarkacron{} (Tab.~\ref{tab:mucca-level2-and-3}). However, despite the increased difficulty, the relative rankings of the methods remain largely consistent with those in the $L=1$ setting. In particular, GroundingREC~\cite{10656642}, CountGD~\cite{DBLP:journals/corr/abs-2407-04619}, and DAVE~\cite{Pelhan_2024_CVPR} continue to exhibit the strongest robustness on the FSC-147 dataset.
%, experiencing less pronounced drops in F1-score compared to density-based approaches such as ZSC and CLIP-Count. 
Furthermore, on the inherently multi-class \benchmarkacron{} dataset, GroundingREC further confirms its state-of-the-art performance, maintaining a clear margin over competing methods even at the highest grid level ($L=3$).

\begin{table*}[!t]
%\caption{\textbf{TODO} TODO. FSC test split.} % test split of fsc, game-2 and game-3
\caption{\textbf{Extended results on the test set of the FSC-147 dataset evaluated at finer spatial granularities.} We report the performance of the evaluated state-of-the-art methods on the distractor test using grid levels $L=2$ (16 patches) and $L=3$ (64 patches). The metrics include the proposed Counting Precision (CntP), Recall (CntR), and F1-Score (CntF1), alongside the GAME(L) metric. Best, second-best, and third-best results are indicated with \gold, \silver, and \bronze, and formatted using \textbf{bold}, \textit{italic}, and \underline{underline}, respectively.}
\label{tab:fsc-test-level2-and-3}
\vspace{3pt}
\newcolumntype{L}{>{\arraybackslash}m{.27\linewidth}}%
\newcolumntype{C}{>{\centering\arraybackslash}X}
%\tiny%
\setlength{\tabcolsep}{0.8pt}
    \begin{tabularx}{0.98\linewidth}{lCCCCCCCC}
    % results on fsc test split game-2, game-3
    \toprule
    & \multicolumn{4}{c}{L=2} & \multicolumn{4}{c}{L=3} \\
    \cmidrule(lr){2-5} \cmidrule(lr){6-9}
    %Model & NMN $\downarrow$ & PCCN $\uparrow$ & Precision $\uparrow$ & Recall $\uparrow$ & F1-Score $\uparrow$ & GAME-1 $\downarrow$ & MAE $\downarrow$ & RMSE $\downarrow$ \\
    %\midrule
    Model & \footnotesize CntP $\uparrow$ & \footnotesize CntR $\uparrow$ & \footnotesize CntF1 $\uparrow$ & \footnotesize GAME(2) $\downarrow$ & \footnotesize CntP $\uparrow$ & \footnotesize CntR $\uparrow$ & \footnotesize CntF1 $\uparrow$ & \footnotesize GAME(3) $\downarrow$ \\
    \midrule
    \midrule
    ZSC~\cite{DBLP:conf/cvpr/XuL0RS23} {\footnotesize (CVPR '23)} & 0.48 & 0.78 & 0.55 & 68.08 & 0.45 & 0.72 & 0.52 & 72.53 \\
    CounTX~\cite{AminiNaieni23} {\footnotesize (BMVC '23)} & 0.64 & 0.70 & 0.61 & 52.59 & 0.60 & 0.66 & 0.57 & 55.86 \\
    CLIP-Count~\cite{DBLP:conf/mm/JiangLC23} {\footnotesize (ACM MM '23)} & 0.47 & 0.72 & 0.53 & 67.89 & 0.45 & 0.67 & 0.50 & 71.54 \\
    VLCounter~\cite{DBLP:conf/aaai/KangMKH24} {\footnotesize (AAAI '24)} & 0.49 & 0.76 & 0.55 & 65.69 & 0.47 & 0.71 & 0.52 & 68.99 \\
    TFPOC~\cite{10483595} {\footnotesize (WACV '24)} & 0.64 & \underline{0.79}\bronze & 0.65 & 48.01 & 0.58 & \underline{0.73}\bronze & 0.60 & 52.56 \\
    DAVE~\cite{Pelhan_2024_CVPR} {\footnotesize (CVPR '24)} & \textit{0.73}\silver & 0.69 & \underline{0.68}\bronze & \textit{39.38}\silver & \textit{0.68}\silver & 0.64 & \underline{0.63}\bronze & \textit{43.13}\silver\mbox{} \\
    PseCo~\cite{DBLP:conf/cvpr/HuangD0ZS24} {\footnotesize (CVPR '24)} & 0.51 & \textit{0.82}\silver & 0.58 & 63.15 & 0.48 & \textit{0.77}\silver & 0.55 & 67.16 \\
    GroundingREC~\cite{10656642} {\footnotesize (CVPR '24)} & \textbf{0.86}\gold & 0.77 & \textbf{0.79}\gold & \textbf{30.80}\gold & \textbf{0.81}\gold & 0.72 & \textbf{0.75}\gold & \textbf{34.29}\gold\mbox{} \\
    UPC~\cite{DBLP:conf/aaai/0018C24} {\footnotesize (AAAI '24)} & 0.66 & 0.67 & 0.64 & 51.80 & 0.57 & 0.58 & 0.56 & 59.95 \\
    CountGD~\cite{DBLP:journals/corr/abs-2407-04619} {\footnotesize (NeurIPS '24)} & \underline{0.71}\bronze & \textbf{0.83}\gold & \textit{0.75}\silver & \underline{40.44}\bronze & \underline{0.67}\bronze & \textbf{0.78}\gold & \textit{0.70}\silver & \underline{44.95}\bronze\mbox{} \\
    \bottomrule
    \end{tabularx}
\end{table*}

\begin{table*}[!t]
%\caption{\textbf{TODO} TODO. FSC val split.} % val split of fsc, game-2 and game-3
\caption{\textbf{Extended results on the validation set of the FSC-147 dataset evaluated at finer spatial granularities.} We report the performance of the evaluated state-of-the-art methods on the distractor test using grid levels $L=2$ (16 patches) and $L=3$ (64 patches). The metrics include the proposed Counting Precision (CntP), Recall (CntR), and F1-Score (CntF1), alongside the GAME(L) metric. Best, second-best, and third-best results are indicated with \gold, \silver, and \bronze, and formatted using \textbf{bold}, \textit{italic}, and \underline{underline}, respectively.}
\label{tab:fsc-val-level2-and-3}
\vspace{3pt}
\newcolumntype{L}{>{\arraybackslash}m{.27\linewidth}}%
\newcolumntype{C}{>{\centering\arraybackslash}X}
%\tiny%
\setlength{\tabcolsep}{0.8pt}
    \begin{tabularx}{0.98\linewidth}{lCCCCCCCC}
    % results on fsc val split game-2, game-3
    \toprule
    & \multicolumn{4}{c}{L=2} & \multicolumn{4}{c}{L=3} \\
    \cmidrule(lr){2-5} \cmidrule(lr){6-9}
    %Model & NMN $\downarrow$ & PCCN $\uparrow$ & Precision $\uparrow$ & Recall $\uparrow$ & F1-Score $\uparrow$ & GAME-1 $\downarrow$ & MAE $\downarrow$ & RMSE $\downarrow$ \\
    %\midrule
    Model & \footnotesize CntP $\uparrow$ & \footnotesize CntR $\uparrow$ & \footnotesize CntF1 $\uparrow$ & \footnotesize GAME(2) $\downarrow$ & \footnotesize CntP $\uparrow$ & \footnotesize CntR $\uparrow$ & \footnotesize CntF1 $\uparrow$ & \footnotesize GAME(3) $\downarrow$ \\
    \midrule
    \midrule
    ZSC~\cite{DBLP:conf/cvpr/XuL0RS23} {\footnotesize (CVPR '23)} & 0.47 & 0.70 & 0.51 & 63.10 & 0.44 & 0.65 & 0.47 & 66.61 \\
    CounTX~\cite{AminiNaieni23} {\footnotesize (BMVC '23)} & 0.61 & 0.64 & 0.56 & 50.76 & 0.57 & 0.60 & 0.52 & 53.42 \\
    CLIP-Count~\cite{DBLP:conf/mm/JiangLC23} {\footnotesize (ACM MM '23)} & 0.46 & 0.65 & 0.49 & 63.50 & 0.43 & 0.60 & 0.46 & 66.43 \\
    VLCounter~\cite{DBLP:conf/aaai/KangMKH24} {\footnotesize (AAAI '24)} & 0.49 & \underline{0.71}\bronze & 0.54 & 58.61 & 0.46 & \underline{0.67}\bronze & 0.50 & 61.33 \\
    TFPOC~\cite{10483595} {\footnotesize (WACV '24)} & 0.66 & 0.69 & 0.60 & 48.10 & 0.59 & 0.63 & \underline{0.55}\bronze & 51.82 \\
    DAVE~\cite{Pelhan_2024_CVPR} {\footnotesize (CVPR '24)} & \underline{0.67}\bronze & 0.61 & 0.61 & \underline{41.55}\bronze & \underline{0.61}\bronze & 0.56 & \underline{0.55}\bronze & \underline{45.17}\bronze\mbox{} \\
    PseCo~\cite{DBLP:conf/cvpr/HuangD0ZS24} {\footnotesize (CVPR '24)} & 0.48 & \underline{0.71}\bronze & 0.52 & 61.72 & 0.45 & 0.66 & 0.49 & 64.64 \\
    GroundingREC~\cite{10656642} {\footnotesize (CVPR '24)} & \textbf{0.86}\gold & \textit{0.76}\silver & \textbf{0.79}\gold & \textbf{29.12}\gold & \textbf{0.81}\gold & \textit{0.71}\silver & \textbf{0.74}\gold & \textbf{32.16}\gold\mbox{} \\
    UPC~\cite{DBLP:conf/aaai/0018C24} {\footnotesize (AAAI '24)} & \underline{0.67}\bronze & 0.65 & \underline{0.64}\bronze & 43.59 & 0.56 & 0.55 & 0.54 & 51.57 \\
    CountGD~\cite{DBLP:journals/corr/abs-2407-04619} {\footnotesize (NeurIPS '24)} & \textit{0.74}\silver & \textbf{0.80}\gold & \textit{0.76}\silver & \textit{34.70}\silver & \textit{0.68}\silver & \textbf{0.74}\gold & \textit{0.71}\silver & \textit{38.73}\silver\mbox{} \\
    \bottomrule
    \end{tabularx}
\end{table*}

\begin{table*}
%\caption{\textbf{TODO} TODO. MUCCA} % val split of fsc, game-2 and game-3
\caption{\textbf{Extended results on our MUCCA dataset evaluated at finer spatial granularities.} We report the performance of the evaluated state-of-the-art methods on the distractor test using grid levels $L=2$ (16 patches) and $L=3$ (64 patches) over intrinsically multi-class scenes. The metrics include the proposed Counting Precision (CntP), Recall (CntR), and F1-Score (CntF1), alongside the GAME(L) metric. Best, second-best, and third-best results are indicated with \gold, \silver, and \bronze, and formatted using \textbf{bold}, \textit{italic}, and \underline{underline}, respectively.}
\label{tab:mucca-level2-and-3}
\vspace{3pt}
\newcolumntype{L}{>{\arraybackslash}m{.27\linewidth}}%
\newcolumntype{C}{>{\centering\arraybackslash}X}
%\tiny%
\setlength{\tabcolsep}{0.8pt}
    \begin{tabularx}{0.98\linewidth}{lCCCCCCCC}
    % results on fsc val split game-2, game-3
    \toprule
    & \multicolumn{4}{c}{L=2} & \multicolumn{4}{c}{L=3} \\
    \cmidrule(lr){2-5} \cmidrule(lr){6-9}
    %Model & NMN $\downarrow$ & PCCN $\uparrow$ & Precision $\uparrow$ & Recall $\uparrow$ & F1-Score $\uparrow$ & GAME-1 $\downarrow$ & MAE $\downarrow$ & RMSE $\downarrow$ \\
    %\midrule
    Model & \footnotesize CntP $\uparrow$ & \footnotesize CntR $\uparrow$ & \footnotesize CntF1 $\uparrow$ & \footnotesize GAME(2) $\downarrow$ & \footnotesize CntP $\uparrow$ & \footnotesize CntR $\uparrow$ & \footnotesize CntF1 $\uparrow$ & \footnotesize GAME(3) $\downarrow$ \\
    \midrule
    \midrule
    ZSC~\cite{DBLP:conf/cvpr/XuL0RS23} {\footnotesize (CVPR '23)} & 0.41 & 0.64 & 0.40 & 38.02 & 0.32 & 0.50 & 0.32 & 41.90 \\
    CounTX~\cite{AminiNaieni23} {\footnotesize (BMVC '23)} & 0.31 & \textbf{0.81}\gold & 0.40 & 39.18 & 0.27 & \textbf{0.68}\gold & 0.34 & 42.48 \\
    CLIP-Count~\cite{DBLP:conf/mm/JiangLC23} {\footnotesize (ACM MM '23)} & 0.40 & \textit{0.76}\silver & \underline{0.46}\bronze & 30.09 & 0.32 & \textit{0.58}\silver & \underline{0.37}\bronze & 34.17 \\
    VLCounter~\cite{DBLP:conf/aaai/KangMKH24} {\footnotesize (AAAI '24)} & \underline{0.42}\bronze & \underline{0.70}\bronze & \underline{0.46}\bronze & 28.67 & \underline{0.34}\bronze & \underline{0.56}\bronze & \textit{0.38}\silver & 32.09 \\
    TFPOC~\cite{10483595} {\footnotesize (WACV '24)} & \underline{0.42}\bronze & 0.62 & 0.43 & \underline{28.01}\bronze & 0.32 & 0.46 & 0.34 & \textit{31.73}\silver\mbox{} \\
    DAVE~\cite{Pelhan_2024_CVPR} {\footnotesize (CVPR '24)} & 0.35 & 0.42 & 0.31 & 29.04 & 0.23 & 0.26 & 0.20 & 32.54 \\
    %\rowcolor{yellow}
    %PseCo & 0.03 & 0.05 & 0.06 & 65.90 & 0.01 & 0.02 & 0.03 & 68.12 \\ % wrong densities
    %PseCo & 0.30 & 0.60 & 0.35 & 41.33 & 0.20 & 0.41 & 0.24 & 46.54 \\
    PseCo~\cite{DBLP:conf/cvpr/HuangD0ZS24} {\footnotesize (CVPR '24)} & 0.30 & 0.60 & 0.35 & 41.72 & 0.20 & 0.41 & 0.24 & 46.96 \\
    GroundingREC~\cite{10656642} {\footnotesize (CVPR '24)} & \textbf{0.63}\gold & 0.69 & \textbf{0.63}\gold & \textbf{14.60}\gold & \textbf{0.51}\gold & 0.55 & \textbf{0.52}\gold & \textbf{18.24}\gold\mbox{} \\
    UPC~\cite{DBLP:conf/aaai/0018C24} {\footnotesize (AAAI '24)} & \textit{0.45}\silver & 0.66 & 0.45 & 31.86 & \textit{0.36}\silver & 0.52 & \underline{0.37}\bronze & 35.56 \\
    CountGD~\cite{DBLP:journals/corr/abs-2407-04619} {\footnotesize (NeurIPS '24)} & 0.41 & 0.69 & \textit{0.48}\silver & \textit{26.05}\silver & 0.32 & 0.52 & \underline{0.37}\bronze & \underline{31.96}\bronze\mbox{} \\
    \bottomrule
    \end{tabularx}
\end{table*}

\section{Implementation Details}
\label{sec:appendix-implementation}
In this section, we provide some additional implementation details expanding Sec.~\ref{sec:sec:experiments-setting}.

\subsection{Converting Outputs to Density Maps} Several methods included in our evaluation do not natively produce a density map, but instead output \emph{instance-centric} predictions such as segmentation masks, bounding boxes, or point locations. Since \benchmarkacron{} requires a unified output interface across methods (i.e., a per-image count together with a spatial prediction that can be interpreted as a density-like map), we adapted these models by converting their outputs into a fixed-resolution \emph{density map representation}.

Concretely, for each predicted instance, we computed a representative point (typically the centroid of a predicted mask or the center of a predicted bounding box), rescaled it to a common canvas of size $384 \times 384$, and then added a small localized kernel at that location. In our implementation, this kernel is a $5 \times 5$ square normalized to sum to one (i.e., each instance contributes unit mass). The final predicted count is obtained by integrating (summing) the resulting density map.

This conversion was applied to the following models: 
\begin{itemize} 
    \item TFPOC: mask outputs are converted to centroids and then to a density map. 
    \item PseCo: post-processed bounding boxes are converted to box centers and then to a density map. 
    \item GroundingREC: predicted boxes are converted to point locations and then rasterized into a density map 
    \item CountGD: filtered detections are converted to a point map (single-pixel impulses at box centers). 
\end{itemize}

\subsection{DAVE Inference Procedure}
We applied minor adjustments to the original DAVE~\cite{Pelhan_2024_CVPR} inference pipeline to make the model compatible with \benchmarkacron{}. Specifically, DAVE is designed under the assumption that the text prompt always refers to an object class actually present in the image. Because of this assumption, the model assigns every prompt to the dominant cluster by selecting the one with the highest CLIP similarity score.
This becomes problematic when the prompt describes a class that does not appear in a (single-class) image. In such negative cases, the assumption is violated, but DAVE still behaves as if it were true: it keeps assigning the prompt to the dominant cluster simply because that cluster yields the highest (although low) CLIP similarity. As a result, the original implementation never triggers the CLIP-based proposal filtering step, which is supposed to remove proposals incompatible with the prompt. All proposals inside the dominant cluster are therefore retained. Since these proposals are always the same for a given image, the final count returned by the model becomes identical for any incorrect prompt. This explains the systematic failures observed in the negative test, where the correct output should instead be zero.

To address this issue, we explicitly activate CLIP-based proposal filtering even when the prompt refers to an absent class. In our modified version, the filtering is restricted to the proposals belonging to the single cluster that actually contains the objects in the image. To determine whether these proposals match the input prompt, we compute a reference CLIP similarity score by feeding the model with the correct (positive) class. This score serves as an upper bound for what a valid match should look like in that specific image. Following the criterion used in the original implementation, a proposal is retained only if its similarity to the input prompt reaches at least 85\% of this reference score. When the prompt describes an absent class, none of the proposals meet this threshold, and all of them are correctly rejected, allowing the model to output zero in the negative test.

\bibliographystyle{cas-model2-names}

\bibliography{cas-refs}


%Author biography without author photo.
%Author biography. Author biography. Author biography.
%Author biography. Author biography. Author biography.
%Author biography. Author biography. Author biography.
%Author biography. Author biography. Author biography.
%Author biography. Author biography. Author biography.
%Author biography. Author biography. Author biography.
%Author biography. Author biography. Author biography.
%Author biography. Author biography. Author biography.
%Author biography. Author biography. Author biography.

%\bio{figs/cas-pic1}
%Author biography with author photo.
%Author biography. Author biography. Author biography.
%Author biography. Author biography. Author biography.
%Author biography. Author biography. Author biography.
%Author biography. Author biography. Author biography.
%Author biography. Author biography. Author biography.
%Author biography. Author biography. Author biography.
%Author biography. Author biography. Author biography.
%Author biography. Author biography. Author biography.
%Author biography. Author biography. Author biography.

%\bio{figs/cas-pic1}
%Author biography with author photo.
%Author biography. Author biography. Author biography.
%Author biography. Author biography. Author biography.
%Author biography. Author biography. Author biography.
%Author biography. Author biography. Author biography.

\end{document}